\pdfoutput=1

\documentclass[11pt]{article}

\usepackage[dvipsnames]{xcolor}

\usepackage[]{acl}

\usepackage{times}
\usepackage{latexsym}

\usepackage[T1]{fontenc}

\usepackage[utf8]{inputenc}

\usepackage{microtype}

\usepackage{enumitem}
\usepackage{amsmath}
\usepackage{bbm}
\usepackage{tabularx}
\usepackage{booktabs}
\usepackage[normalem]{ulem}
\usepackage{graphicx}
\usepackage{caption}
\usepackage{subcaption}
\usepackage{multirow}
\usepackage{cleveref}

\usepackage{longtable}

\newcommand{\MODEL}{\textsc{FactPEGASUS}}

\title{ \MODEL{}: Factuality-Aware Pre-training and Fine-tuning for Abstractive Summarization}

\author{David Wan \and Mohit Bansal \\
University of North Carolina at Chapel Hill \\
\texttt{\{davidwan,mbansal\}@cs.unc.edu}
}

\begin{document}
\maketitle
\begin{abstract}

We present \MODEL{}, an abstractive summarization model that addresses the problem of factuality during pre-training and fine-tuning:
(1) We augment the sentence selection strategy of PEGASUS's \cite{zhang2019pegasus} pre-training objective to create pseudo-summaries that are both important and factual;
(2) We introduce three complementary components for fine-tuning.
The \textit{corrector} removes hallucinations present in the reference summary,
the \textit{contrastor}
uses contrastive learning to
better differentiate nonfactual summaries from factual ones,
and the \textit{connector} bridges the gap between the pre-training and fine-tuning for better transfer of knowledge.
Experiments on three downstream
tasks demonstrate that \MODEL{} substantially improves
factuality evaluated by multiple automatic metrics and humans.
Our thorough analysis suggests that \MODEL{}
is more factual than using the original pre-training objective in zero-shot and few-shot settings,
retains factual behavior more robustly than strong baselines,
and does not rely entirely on becoming more extractive to improve factuality.\footnote{Our code and data are publicly available at: \url{https://github.com/meetdavidwan/factpegasus}.}

\end{abstract}

\section{Introduction}

Abstractive summarization aims at generating short summaries
that capture the essentials of a long document.
Research in this challenging task has made
significant progress
with the help of large pre-trained models \cite{lewis-etal-2020-bart, 2020t5,zhang2019pegasus}. However, current models suffer from the crucial problem of hallucinations \cite{maynez-etal-2020-faithfulness},
where a summary contains facts or entities not present in the original document. Such unfaithful generation raises the question of whether the models can be trustworthy and used safely for real-world applications.
To tackle this problem, many approaches propose post-processing models \cite{chen-etal-2021-improving, dong-etal-2020-multi, liu-liu-2021-simcls}, but such methods are often constrained by external resources to train additional correction or selection models. An alternative line of works focuses on learning factuality directly during fine-tuning by filtering nonfactual training data \cite{goyal-durrett-2021-annotating, nan-etal-2021-entity} or, most recently, incorporating contrastive learning \cite{cao-wang-2021-cliff} to encourage generating faithful summaries.

In this work, we propose \MODEL{}, a model that addresses the problem of hallucinations for abstractive summarization holistically, by incorporating factuality into the whole training pipeline: We tackle the lack of factuality objective in pre-training and the presence of hallucinations in the downstream dataset during fine-tuning.
Current pre-training objectives focus on improving the quality of the generated output in the downstream tasks but often overlook the factuality aspect.
Thus, we explore incorporating \textbf{factuality into
the pre-training objective} of PEGASUS \cite{zhang2019pegasus} (a state-of-the-art abstractive summarization model).
The original objective, gap sentence generation (GSG), transforms any text into a pseudo-summarization dataset by selecting important sentences using ROUGE \cite{lin-2004-rouge} as output summaries.
We explore strategies for combining ROUGE and the factuality metric FactCC \cite{kryscinski-etal-2020-evaluating} as the selection criteria, so that the model learns to generate sentences that cover the most important information of the input document as well as remain faithful to it.

Next, we propose three complementary modules that further address factuality problems during fine-tuning:
(1) \textbf{Corrector} that removes hallucinations existing in reference summaries,  
allowing training on the full training set without learning unfaithful behaviors;
(2) \textbf{Contrastor} that encourages the model to better differentiate factual summaries from nonfactual ones by paying attention to the document using contrastive learning;
(3) \textbf{Connector},
a special mask-token fine-tuning technique enabled by the GSG-style objective, that simulates the pre-training task
during fine-tuning
by inserting the mask token into the input document  so that the pre-trained model can adapt its knowledge of generating factual summaries directly to the downstream tasks.
The connector, corrector, and contrastor address the input, output, and training objective of the downstream task, respectively, and the combination of the components reduces potential confounding problems that cannot be addressed by a single module.
We show that the full model improves three factuality metrics, the token and sentence error of DEP Entail \cite{goyal-durrett-2021-annotating} and FactCC, on the downstream datasets of XSum \cite{narayan-etal-2018-dont}, WikiHow \cite{koupaee2018wikihow}, and Gigaword \cite{rush-etal-2015-neural}.
Most notably, \MODEL{} outperforms existing factuality-aware summarization models by more than $40\%$ and $34\%$ on XSum for token error and FactCC, respectively.
Ablation studies show the usefulness of each of our fine-tuning components as well as the additive gain of combining our complementary modules,
and human evaluation confirms that \MODEL{} generates significantly more factual summaries over strong baselines.

Finally, we perform a detailed analysis of \MODEL{}, which points to several important observations regarding learning and maintaining factuality: (1)
\textbf{Zero-shot} setting demonstrates the utility of our factuality-aware pre-training objective, as our model outperforms PEGASUS (which uses the original objective) on all three factuality metrics when evaluated directly on the downstream task
without any supervised training data.
\textbf{Few-shot} experiment indicates that even a small number of nonfactual examples can have a strong negative impact on factuality and can nullify much of the gain from factuality pre-training, highlighting the importance of ensuring factuality during fine-tuning.
(2) \textbf{Factuality dynamics}
\cite{goyal2021training} further shows that \MODEL{} exhibits a lesser degree of factuality degradation than what is observed for BART-base.
(3) \textbf{Factuality vs abstractiveness tradeoff curve} reveals that \MODEL{} effectively improves factuality by not simply relying on the increase in extractiveness.

To summarize, our contributions are as follows:
\begin{enumerate}[leftmargin=*]
\setlength{\itemsep}{-1pt}
     \item We propose a factuality-aware pre-training objective for abstractive summarization and study the effect of different sentence selection strategies on downstream factuality.
     \item We introduce three complementary components for improving factuality during fine-tuning that correct hallucinations present in the training set, discourage unfaithful generation during training, and bridge the gap between pre-training and fine-tuning.
     The full model consistently achieves better factuality scores than strong baselines
     on three downstream abstractive summarization tasks, confirmed by human evaluation.
     \item
    We conduct thorough factuality analysis and show that \MODEL{} generates more factual summaries with no or little supervision, slows down factuality degradation observed for current models, and improves factuality not by becoming more extractive.
\end{enumerate}

\section{Related Work}

\paragraph{Pre-training Objective for Generation Tasks.}
Transformer-based models have achieved state-of-the-art performance for abstractive summarization \cite{devlin-etal-2019-bert, lewis-etal-2020-bart, 2020t5, zhang2019pegasus}. Many such pre-trained models study the effect of useful pre-training objectives, often in the form of masking certain parts of the input. BART \cite{lewis-etal-2020-bart} randomly masks spans of tokens in the text as input and asks the model to reconstruct the original text.
Our work builds on the success of PEGASUS's \cite{zhang2019pegasus} pre-training objective that closely resembles the downstream summarization task.
Their objective selects sentences that best represent the document as the output summary, and masks out the selected sentences in the original text as the input document. We explore various sentence selection strategies to encourage the model to generate summaries that cover the most important information of the document and also remain faithful to it.

\paragraph{Improving Factuality for Summarization.} Recent models can achieve highly fluent and coherent abstractive summaries, yet the generated summaries often contain factual errors \cite{falke-etal-2019-ranking, maynez-etal-2020-faithfulness}. Several approaches have addressed this problem, which can be roughly categorized into two types. The first approach proposes post-processing models, that either removes hallucinations in the generated summaries \cite{cao-etal-2020-factual, dong-etal-2020-multi}, or selects the most factual candidate during beam search \cite{chen-etal-2021-improving}. This approach often requires training additional models and external resources.
In an attempt to improve factuality in an end-to-end fashion,  \citet{nan-etal-2021-entity} and \citet{goyal-durrett-2021-annotating} explore a useful method of
removing nonfactual examples during training, but this only allows the model to be trained on a small portion of the training data. 

Recently, contrastive learning \cite[CL]{1467314} has started to gain traction for improving factuality. Popular for representation learning, CL has had great success for vision tasks \cite{chen2020simple}
and has also been successfully applied to summarization, where \citet{liu-liu-2021-simcls} improves summary quality by differentiating high-quality summaries from the lower-quality ones.
\citet{cao-wang-2021-cliff} extend this idea to improve factuality with various approaches to generate hallucinated summaries as negative examples, showing consistent improvement over existing methods. We similarly incorporate CL as an additional training objective, but we differ from previous works
in the choice of anchor and positive sample.
Inspired by \citet{lee2021contrastive}, who use encoder and decoder output as candidates for CL across multiple text generation tasks, we extend this idea to factuality, i.e., instead of performing CL only between summaries, we
perform CL between the document and the summary.
This setup encourages the model to generate a faithful summary that pays attention to the document, i.e., the definition of faithfulness.

\begin{figure*}[t]
    \centering
    \begin{subfigure}{.5\textwidth}
    \centering
    \includegraphics[width=\linewidth]{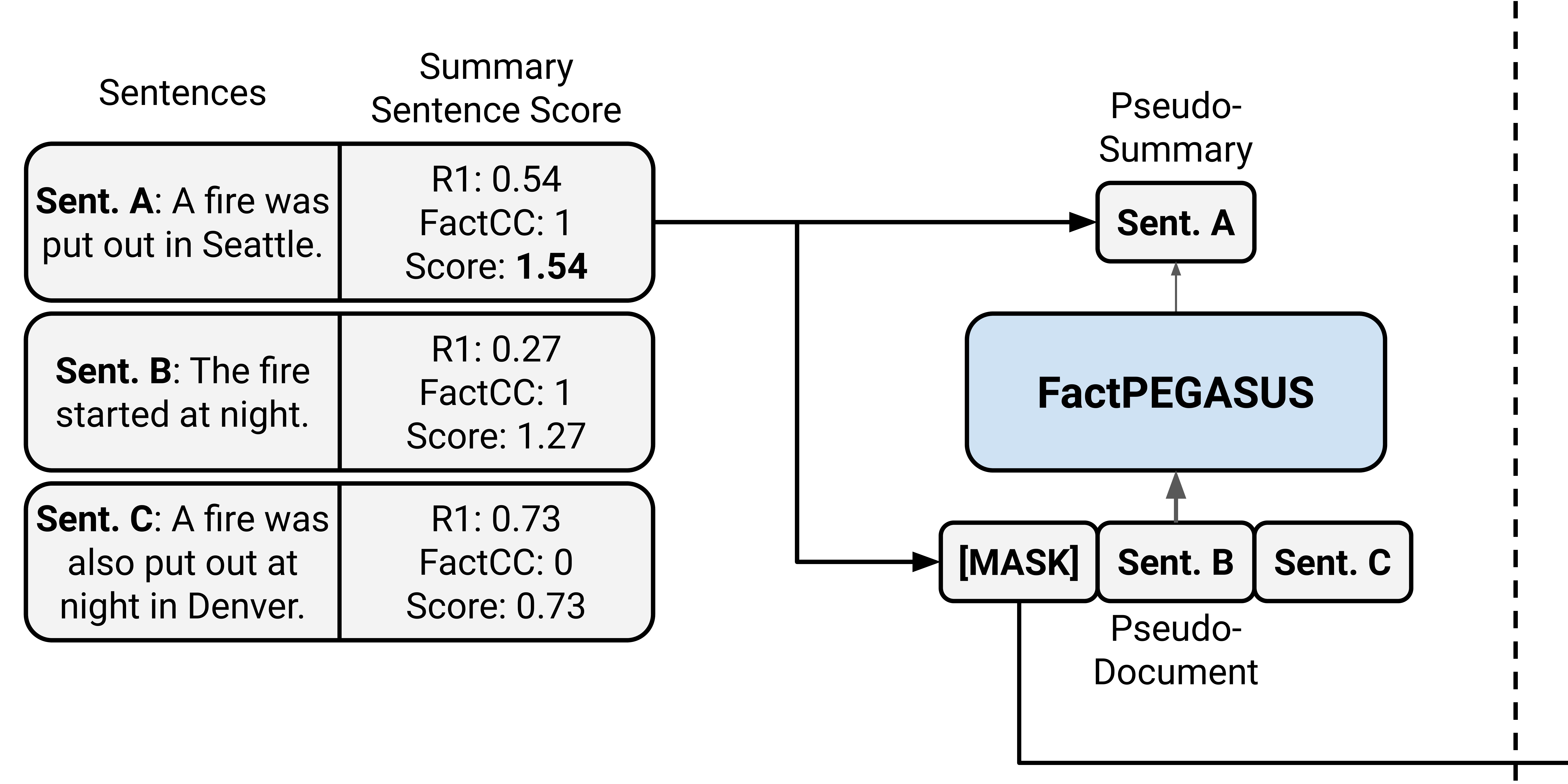}
    \caption{Pre-training}
    \label{fig:model_pretraining}
    \end{subfigure}%
    \begin{subfigure}{.5\textwidth}
    \centering
    \includegraphics[width=\textwidth]{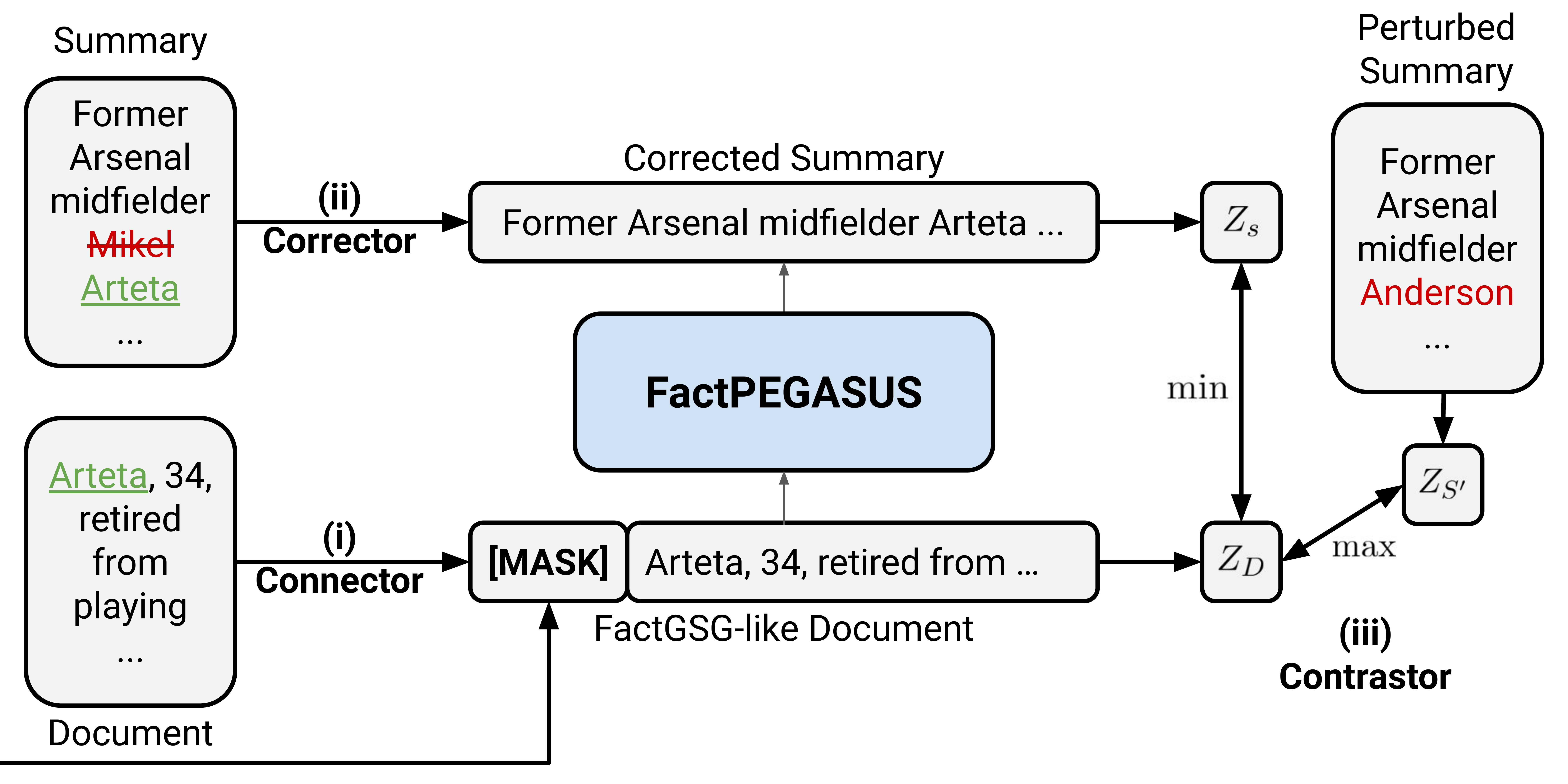}
    \caption{Fine-tuning}
    \label{fig:model_finetuning}
    \end{subfigure}
    \vspace{-3pt}
    \caption{Illustration of \MODEL{}. For pre-training (a), we use the factGSG objective introduced in Section \ref{sec:pretraining} that transforms a text document into a pseudo-summarization dataset. We select the pseudo-summary using the combination of ROUGE and FactCC.
    Here, sentence A is selected as the pseudo-summary, and we mask this sentence in the original text to create the pseudo-document.
    During fine-tuning (b), the connector (i) simulates the factGSG task by appending the same mask token used in (a) to the input document, so that we have the same setup in both training stages. Then, corrector (ii) removes hallucinations (highlighted in \textcolor{red}{red}) from the summary. Finally, contrastive learning in (iii) encourages the model to prefer the corrected summary over the perturbed summary.
    }
    \vspace{-6pt}
    \label{fig:model_figure}
\end{figure*}
\section{\MODEL{}}
We describe our training procedure consisting of pre-training with a factuality-aware objective (Section~\ref{sec:pretraining}) and fine-tuning with three complementary modules for improving factuality (Section ~\ref{sec:finetuning}).

\subsection{Factuality-Aware Pre-training}\label{sec:pretraining}

Recent exploration of good pre-training objectives for abstractive summarization aims at achieving high quality on downstream tasks, often in terms of ROUGE. However, few have analyzed the effect of pre-training objective on factuality.
We focus on incorporating this aspect into the pre-training objective of PEGASUS, gap sentence generation (\textbf{GSG}), since PEGASUS achieves state-of-the-art performance on the downstream abstractive summarization tasks.
The GSG objective transforms text documents into a pseudo-summarization dataset by selecting important sentences as the output summary, which are subsequently masked out in the original text. The best strategy determines the importance by calculating ROUGE-1 between each chosen sentence and the rest of the document.
While the original strategy selects sentences that contain the most unigram overlap, there is no guarantee that the selected sentences are faithful to the rest of the document. We provide an illustrative example in \autoref{fig:model_pretraining}, where the original objective selects sentence C due to its high ROUGE-1 score. However, this sentence is not a faithful summary to the rest of the document as the other sentences concern with the fire in Seattle while only sentence C talks about the fire in Denver.

To address this problem, we extend this objective, which we call factual GSG (\textbf{factGSG}), where we additionally measure the importance of the sentences according to factuality. 
We use FactCC \cite{kryscinski-etal-2020-evaluating} as the factuality criteria when selecting the summary sentences, as it correlates highly with human factuality judgment \cite{pagnoni-etal-2021-understanding} and is relatively fast to compute. FactCC produces a binary prediction where a score of 1 indicates that the selected sentence is consistent with the rest of the document.
Another change in factGSG is the choice of gap sentence ratio, which determines the percentage of sentences in the text that will be selected as the summary. Instead of selecting $30\%$ of the text document as output summary, we only select one sentence, as selecting more sentences will inevitably increase the possibility of hallucinations.

Formally, given a document $D$ of $n$ sentences, $D = \{x_1, x_2, ..., x_n\}$, we select the top-scoring sentence as the output summary, where the score of each sentence $x_i$ is calculated by:
\[s_i =  rouge(x_i, D \setminus \{x_i\}) + FactCC(x_i, D \setminus \{ x_i \}) \]

Going back to the example in \autoref{fig:model_pretraining}, FactCC assigns a score of 0 to the nonfactual sentence C because the fire in Denver is not entailed by the other sentences. This results in sentence A scoring higher than the nonfactual sentence, and thus overcomes the problem in the original objective.

\subsection{Factuality-Aware Fine-tuning}\label{sec:finetuning}

Although the typical approach of updating all the model's parameters during fine-tuning adapts well to the downstream task, the model suffers from imitative falsehood \cite{lin2021truthfulqa}: The model learns to generate similar hallucinations present in the downstream dataset, and even completely forgets its factual behaviors learned during pre-training. This is especially problematic for datasets like XSum that contains hallucinations on $70\%$ of the summaries \cite{maynez-etal-2020-faithfulness}.

To this end, we present three complementary fine-tuning modules, illustrated in
\autoref{fig:model_finetuning}. Each component addresses different parts of the downstream task and collaboratively ensures factuality throughout the fine-tuning stage.

\subsubsection{Connector} \label{sec:mask_finetuning}
The GSG objective enables faster and better adaptation during fine-tuning by simulating the downstream task \cite{zhang2019pegasus}. However, there still exists a gap between pre-training and fine-tuning: GSG is a masked sentence prediction task, but downstream summarization does not make use of the mask token. Thus, we simply insert the mask token into the input document of the downstream dataset, so as to simulate what the model expects during pre-training. This can be seen as a form of prompting, which helps us to elicit the factuality knowledge of the pre-trained models.
We insert the mask token between sentences, and the best position is determined by
evaluating the summarization performance on the validation set.
We report the best position of the mask token and discuss the similarity to prompting in \autoref{sec:mask_token_position}. 

\newcommand{\ent}[2]{[#1]\textsubscript{#2}}
\newcommand{\factent}[1]{\underline{\textcolor{ForestGreen}{#1}}}
\newcommand{\halent}[1]{\textcolor{red}{#1}}

\begin{figure*}[ht]
    \centering
    \normalfont
    \small
    \begin{tabularx}{\textwidth}{l X}
        \toprule
        Document & \ent{\factent{Arteta}}{ent}, \ent{34}{number}, retired from playing at \ent{the end of last season}{date} ... \ent{\factent{Arteta}}{ent} was seen crying after his final \ent{\factent{Arsenal}}{ent} match...
        \ent{Guardiola}{ent}'s first game since succeeding \ent{Manuel Pellegrini}{ent} ... \\
        Summary &  Former \ent{\factent{Arsenal}}{ent} midfielder \ent{\halent{Mikel} \factent{Arteta}}{ent} has taken up a coaching role at \ent{\halent{Manchester City}}{ent}.\\
        \midrule
        \multicolumn{2}{c}{\textbf{Corrector}}\\
        Replace & Former \factent{Arsenal} midfielder \factent{Arteta} has taken up a coaching role at \halent{Manchester City}. \\
        Remove & Former \factent{Arsenal} midfielder \sout{Mikel Arteta} has taken up a coaching role \sout{at Manchester City}. \\
        Combined & Former \factent{Arsenal} midfielder \underline{\textcolor{ForestGreen}{Arteta}} has taken up a coaching role \sout{at Manchester City}. \\
        \midrule
        \multicolumn{2}{c}{\textbf{Contrastor}}\\
        Intrinsic &  Former Arsenal midfielder \textit{Manuel Pellegrini} has taken up a coaching role. \\
        Extrinsic &  Former Arsenal midfielder \textit{Wenger} has taken up a coaching role. \\
        
        \bottomrule
    \end{tabularx}
    \vspace{-3pt}
    \caption{Example output using different strategies of corrector and contrastor. The first two rows show the original document and summary with highlighted entities and their respective labels (date, number, ent). We mark hallucinated entities in the summaries with \halent{red}, factual entities in document and summary with \factent{green and underlined}, and removed entities by the corrector with a \sout{strikethrough}. Perturbed entities by the contrastor are \textit{italicized}.}
    \vspace{-6pt}
    \label{fig:corrector_contrastor_example}
\end{figure*}

\subsubsection{Corrector}\label{sec:corrector}
The corrector removes hallucinations
in the reference summaries
so that such examples can be used during training without contributing to the problem of imitative falsehood.
We consider summary entities as hallucinating if the text cannot be matched to one of the document entities.
We propose three approaches with varying degrees of aggressiveness w.r.t. the removal of hallucinations and the possibility of generating ungrammatical sentences.

\textbf{Replace:} Upon qualitative analysis, we discover that some hallucinated entities in the summary are partially present in the documents. The most prominent example is the use of names, where the summary contains the full name of the person while only the first or last name is mentioned in the document, as shown in \autoref{fig:corrector_contrastor_example}.
Given such observation, we propose a method to find a similar entity with the same NER label in the document and use that to replace the original hallucinated entity in the summary. Although this approach cannot correct hallucinations where similar entities are missing in the document, grammaticality is ensured.

\textbf{Remove:} A more aggressive approach is to remove the hallucinated entities in the training examples.
The intuition is that it is often better to not say anything than to say something wrong.
We mitigate the problem of creating ungrammatical sentences by removing related words to the removed entities determined by dependency arcs.

\textbf{Combined:} As a middle ground that ensures no hallucinations are present in the reference summaries while being grammatical when possible, we first replace all possible entities and then apply the remove strategy on the remaining ones.

We refer the readers to  Appendix-\ref{sec:corrector_details} for the details about hallucination detection, as well as the algorithm and discussion of grammatically for the remove method.
\subsubsection{Contrastor}\label{sec:contrastor}
To better distinguish factual summaries from nonfactual ones, we next introduce a contrastive learning objective that encourages the model to prefer factual summaries given the context of the document.
We use the document $D_i$ as the anchor and only consider the reference summary $S_i$ as the positive sample.
Then, we create a set of nonfactual summaries $N_i$ to form negative pairs following \citet{kryscinski-etal-2020-evaluating}, where we replace factual entities with random entities of the same named entity labels. We experiment with two variants simulating either extrinsic and intrinsic hallucinations. As formulated in \citet{maynez-etal-2020-faithfulness}, extrinsic hallucinations refer to entities that are present in the summary but not in the document, whereas intrinsic hallucinations are those that are present in the document but contain inaccurate information or are misplaced. See Appendix~\ref{sec:contrastor_details} for more details.

We stress that we perform contrastive learning between the document and the summary, similar to \citet{lee2021contrastive}, instead of between summaries \cite{cao-wang-2021-cliff}, as it follows closer to the definition of faithfulness - the summary should be generated within the context of the document.

We use the NT-Xent loss \cite{chen2020simple}:
\[ l_{D_i, S_i} = - \log{ \frac{ \exp{\text{sim}(z_{D_i}, z_{S_i}) / \tau } } { \sum_{S_j \in N_i \cup \{S_i\}} \exp{\text{sim}(z_{D_i}, z_{S_j}) / \tau } } } \]
where $z_{D_i}$, $z_{S_i}$ and $z_{S_j}$ are representation for $D_i$, $S_i$ and $S_j$, respectively. We generate $z_{D}$ and $z_{S}$ by performing mean pooling over the last hidden layer of the encoder and decoder output, respectively. $\text{sim}( \cdot ,  \cdot )$ is the cosine similarity between the representations, and $\tau$ is the temperature parameter.

The final loss is calculated by the sum of the cross-entropy loss $L_{CE}$ and the contrastive loss: $L = L_{CE} + \lambda L_{CL}$, where $\lambda$ is a scalar.

\section{Experimental Setup}
We describe our experimental setup, and refer to \autoref{sec:experiment_details} for more details.

\subsection{Datasets and Evaluation Metrics}

\begin{figure}[t]
    \centering
    \includegraphics[width=0.99\columnwidth, keepaspectratio]{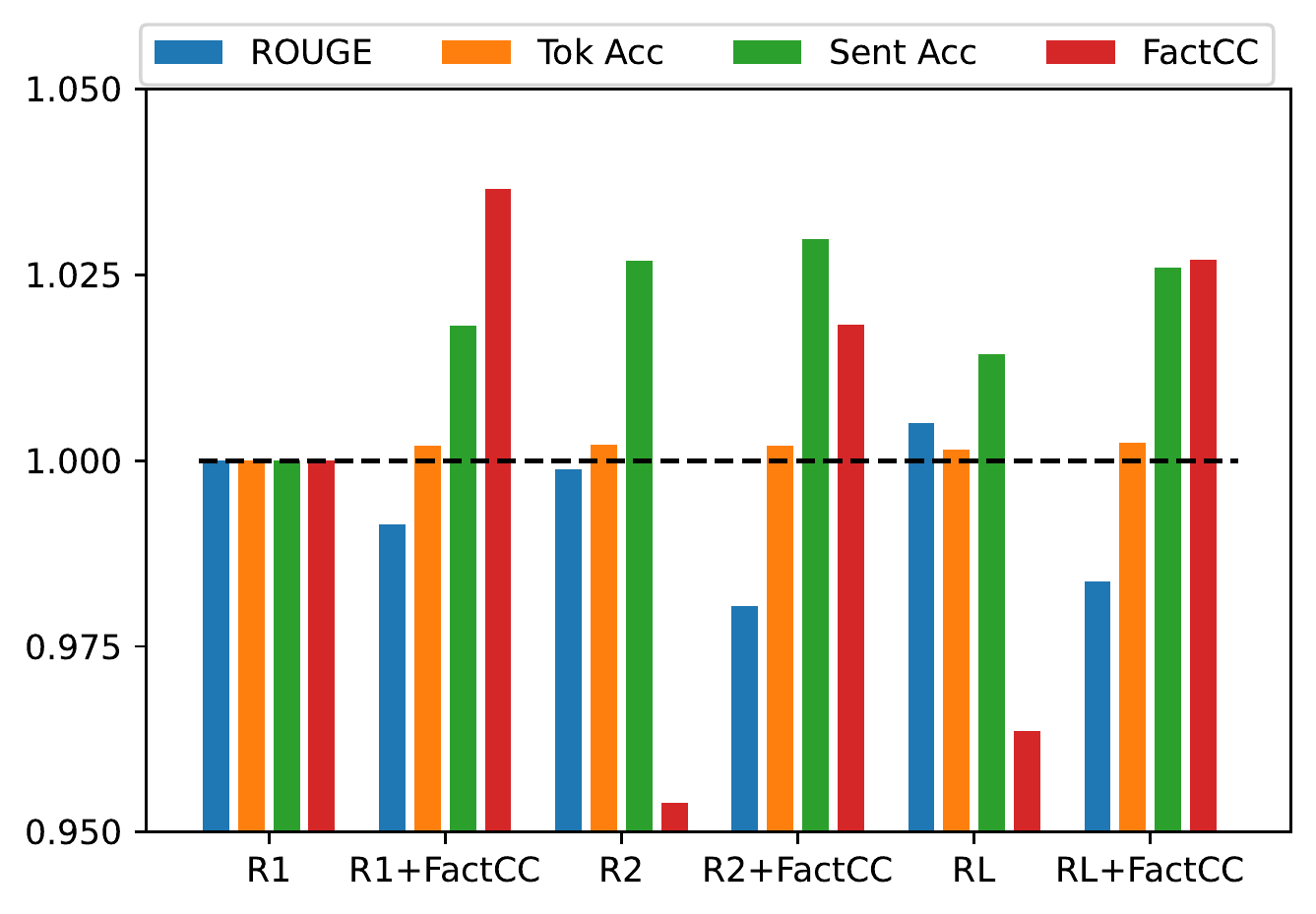}
    \vspace{-9pt}
    \caption{Relative effect using different sentence selection criteria on XSum. Adding FactCC to criteria consistently improves factuality. Full result in \autoref{tab:gsg_metric_full}.
    }
    \vspace{-6pt}
    \label{fig:gsg_metric}
\end{figure}

We pre-train on the C4 dataset \cite{2020t5}, and evaluate our pre-trained model on three downstream abstractive summarization datasets: XSum \cite{narayan-etal-2018-dont}, WikiHow \cite{koupaee2018wikihow}, and Gigaword \cite{rush-etal-2015-neural}. XSum is the primary dataset for analysis unless otherwise stated, as most of the factuality works for abstractive summarization evaluate on this dataset. Dataset details are presented in Appendix~\ref{sec:dataset_details}.

We report ROUGE-L \cite{lin-2004-rouge} to evaluate our generated summaries against the reference. However, we note that this method is not ideal given the presence of hallucinations in the reference summaries \cite{chen-etal-2021-improving,maynez-etal-2020-faithfulness}: If a more factual model does not produce such hallucinations, the output is scored lower than those that contain the same hallucinations found in the reference.

To evaluate factuality, there have been many proposed automatic metrics \cite{durmus-etal-2020-feqa,wang-etal-2020-asking,scialom-etal-2021-questeval}. We report FactCC \cite{kryscinski-etal-2020-evaluating} and DEP-Entail \cite{goyal-durrett-2021-annotating}, as they are highly correlated with human judgment of factuality \cite{pagnoni-etal-2021-understanding}. For DEP-Entail, we report the token-level and sentence-level error. For FactCC, since the model has been trained to evaluate on single sentences, we calculate the average score across all sentences for each summary.

To confirm our observation, we conduct human evaluation asking Amazon Mechanical Turk\footnote{\url{https://www.mturk.com/}} (AMT) to judge the factuality and informativeness of the summaries. We randomly select 100 documents and ask the annotators to check whether each of the generated summaries is factual and informative.
\autoref{sec:human_eval_detail} provides more details.

\subsection{Pre-training and Fine-tuning Setup}
For pre-training, we use BART-base's architecture with PEGASUS's SentencePiece \cite{kudo-2018-subword} unigram model tokenizer.
We first determine the best sentence selection criteria by experimenting with selection criteria that use ROUGE-1, ROUGE-2, and ROUGE-L, as well as combining each with FactCC. To save computation \cite{lewis-etal-2020-bart, zhang2019pegasus, 2020t5}, we pre-train these models on a smaller dataset and fewer training steps.
We report the effect of the selection criteria using the normalized ROUGE score
and factuality scores over the model that uses ROUGE-1 as the selection criteria. We take the complement of token error and sentence error as token accuracy and sentence accuracy, respectively, to present all metrics where higher is better. Details of pre-training are shown in Appendix~\ref{sec:pretraining_detail}.

Finally, We evaluate our pre-trained model
on the three downstream tasks.
As baselines, we compare our model to BART-base and PEGASUS*, our variant of the PEGASUS-base as there is no publicly available checkpoint. We train PEGASUS* by using the original sentence selection metric (ROUGE-1), and observe higher ROUGE scores on XSum and WikiHow than the ones reported in the original paper. We also compare \MODEL{} to two summarization models optimized for factuality. DAE \cite{goyal-durrett-2021-annotating} uses DEP-Entail to mask out the nonfactual tokens during training, and CLIFF \cite{cao-wang-2021-cliff} uses contrastive learning between the reference summaries and automatically generated nonfactual summaries. We apply both methods to BART-base. Details are described in Appendix~\ref{sec:finetuning_detail}.

\section{Result}

\subsection{Pre-training Sentence Selection Results}\label{sec:pretraining_result}

\autoref{fig:gsg_metric} shows the effect of different sentence selection criteria. Adding FactCC to all three ROUGE-only criteria consistently improves all factuality metrics at the cost of a small decrease in quality. Overall, the selection strategy of combining ROUGE-1 and FactCC achieves the highest FactCC score out of all strategies while maintaining the smallest relative drop in ROUGE.

\begin{table}[t]
    \centering
    \small
    \resizebox{\columnwidth}{!}{
    \begin{tabular}{c c c c c c  }
    \toprule
    Dataset & Model & RL & tok err$\downarrow$ & sent err$\downarrow$ & FactCC \\
      \midrule
    \multirow{5}{*}{XS} & BART-base & \textbf{33.78}&  12.38$\phantom{*}$ & 60.70$\phantom{*}$ & 23.99 \\
    & PEGASUS* & 33.17 & 12.33$\phantom{*}$ & 60.01$\phantom{*}$ & 24.14 \\
    \cmidrule{2-6}
    & DAE & 31.78 & $\phantom{0}$4.79* & 35.52* & 25.43 \\
    & CLIFF & 31.40 & 10.36$\phantom{*}$ & 53.14$\phantom{*}$ & 23.77 \\
    & \MODEL{} & 31.17 & \textbf{6.07} & \textbf{38.66}$\phantom{*}$ & \textbf{34.32}  \\
    \midrule
    \multirow{5}{*}{WH} & BART-base & 31.81 & $\phantom{0}$8.99$\phantom{*}$ & 45.77$\phantom{*}$ & 99.09 \\
    & PEGASUS* & 30.30 & $\phantom{0}$9.77$\phantom{*}$ & 47.28$\phantom{*}$ & 98.83 \\
    \cmidrule{2-6}
    & DAE & 31.66 & 4.91* & 34.45* & 98.87 \\
    & CLIFF & \textbf{33.82} & 13.74$\phantom{*}$ & 57.42$\phantom{*}$ & 99.18 \\
    & \MODEL{}  & 29.33 & \textbf{7.86} & \textbf{42.40}$\phantom{*}$ & \textbf{99.41}  \\
    \midrule
    \multirow{5}{*}{GW} & BART-base & \textbf{35.11} & $\phantom{0}$2.29$\phantom{*}$ & 19.68$\phantom{*}$ & 55.66   \\
    & PEGASUS* &  34.74 & $\phantom{0}$2.84$\phantom{*}$ & 22.66$\phantom{*}$ & 56.43 \\
    \cmidrule{2-6}
    & DAE & 35.57 & $\phantom{0}$0.58* & $\phantom{0}$7.54* & 59.61 \\
    & CLIFF & 34.89 & $\phantom{0}$\textbf{1.72}$\phantom{*}$ & \textbf{18.45}$\phantom{*}$ & 58.53 \\
    & \MODEL{}  & 34.23 & $\phantom{0}$2.30$\phantom{*}$ & 19.32$\phantom{*}$ & \textbf{60.02} \\
    \bottomrule
    \end{tabular}
    }
    \vspace{-3pt}
    \caption{Fine-tuning results on the XSum (XS), WikiHow (WH), and Gigaword (GW) dataset. \MODEL{} consistently improves factuality metrics for all datasets over the two baseline models, and outperforms existing factuality models on FactCC. The token error and sentence error achieved by DAE (marked with *) is not a fair comparison, because the model optimizes the metric during training.
    }
    \vspace{-6pt}
    \label{tab:final_base}
\end{table}
\begin{table}[t]
    \centering
    \small
    \begin{tabular}{l c c c}
    \toprule
     Model & Factuality & Informativeness \\ 
    \midrule
    BART-base &  24.67 & 61.33 \\ 
    PEGASUS* & 27.33 & 58.33 \\
    DAE & 31.99 & 61.66 \\
    CLIFF & 29.33 & \textbf{62.99} \\
    \MODEL{} & \textbf{39.66} & 58.67 \\ 
    \bottomrule
    \end{tabular}
    \vspace{-3pt}
    \caption{Human evaluation results on XSum. Our model is statistically significantly better ($p<0.05$) than BART-base, PEGASUS*, and CLIFF, and moderately significantly better than DAE ($p=0.055$). There is no statistical significance between the informativeness of \MODEL{} and other models ($p>0.15$). }
    \label{tab:human_eval}
    \vspace{-6pt}
\end{table}
\subsection{Fine-tuning Results}
We present our full result on the three downstream tasks in \autoref{tab:final_base}. While the two baseline models achieve similar factuality scores, \MODEL{} consistently improves factuality over the two baselines on all three datasets. The largest improvement can be seen for the XSum dataset, where \MODEL{}, compared to BART-base, 
lowers the token error and sentence error by $51\%$ and $36\%$, respectively, and increases FactCC by $43\%$
\footnote{We also experimented with a more aggressive corrector that can achieve more than $50\%$ increase in FactCC and $59\%$ improvement on sentence error on XSum, but this variant can hurt informativeness. Hence, the results can be tuned depending on the desired tradeoff between factuality and informativeness on the downstream task at hand. }
. The same trend but to a lesser degree can also be observed for WikiHow and Gigaword, most notably a 3-point decrease in sentence error for WikiHow and a 2-point increase in FactCC for Gigaword.

Compared to factuality-aware models, \MODEL{} achieves the highest FactCC on all tasks. Notably, \MODEL{} outperforms DAE by $34\%$ on XSum. In terms of DEP-Entail, \MODEL{} outperforms CLIFF on XSum and WikiHow. We note that DAE is trained using the DEP-Entail metric and thus is not a fair comparison.

We note that the ROUGE-L scores for \MODEL{} are lower than both baseline models by about 2 points,
but we stress that our increase in FactCC is substantially larger than the decrease in ROUGE-L for XSum and Gigaword.
The negative relationship between factuality metrics and ROUGE is also reported in prior works \cite{chen-etal-2021-improving,kryscinski-etal-2019-neural}. For example, fine-tuning BART on a subset of XSum \cite{goyal-durrett-2021-annotating} improves factuality at the cost of a 6-point drop in ROUGE-L\footnote{The result is  reported in \citet{cao-wang-2021-cliff}.}, which is triple the amount of decrease observed for our model.

\textbf{Human Evaluation} results are shown in \autoref{tab:human_eval}. The result agrees with our observation on automatic factuality metrics, as \MODEL{} produces significantly more factual summaries than the BART-base, and PEGASUS*, and CLIFF. We achieve moderately significantly better summaries ($p=0.055$) than DAE. Although, \MODEL{} achieves low informativeness, we find no statistical significant difference between our model and other models ($p>0.15$).

\begin{table}[t]
    \centering
    \resizebox{\columnwidth}{!}{
    \begin{tabular}{l c c c c  }
    \toprule
     Model & RL & tok err$\downarrow$ & sent err$\downarrow$ & FactCC \\
    \midrule
    factGSG  &   \textbf{32.99} & 12.31 & 59.30 & 24.94 \\
    \midrule
    + corrector replace &  32.48 & 10.57 & 55.05 &  25.06 \\
    + corrector remove &  30.37  & $\phantom{0}$6.44 & 39.89 &  \textbf{35.77} \\
    + corrector combined &  31.19 & $\phantom{0}$6.10 & 38.96 & 33.79 \\
    \midrule
    + contrastor intrinsic & 32.14 & 11.46 & 57.61  & 25.26 \\
    + contrastor extrinsic & 32.54  & 11.95 & 59.10 & 25.07 \\
    \midrule
    + contrastor + corrector & 31.17 & $\phantom{0}$6.08 & 38.92 & 34.17 \\
    \midrule
    \MODEL{} & 31.17 &  \textbf{$\phantom{0}$6.07} & \textbf{38.66} & 34.32 \\
    \bottomrule
    \end{tabular}
    }
    \vspace{-3pt}
    \caption{Fine-tuning ablation on XSum. We present our pre-trained model factGSG fine-tuned without any of our proposed components, and adding different strategies of corrector and contrastor. We then combine the best of the two modules (corrector combined and contrastor intrinsic), and finally add the connector to form the final model, which we copy from \autoref{tab:final_base}.}
    \label{tab:ablation}
    \vspace{-3pt}
\end{table}

\begin{table}[t]
    \centering
    \small
    \begin{tabular}{c c c c c }
    \toprule
    Model & RL & tok err$\downarrow$ & sent err$\downarrow$ & FactCC \\
    \midrule
    GSG+mask & 23.49 & 9.04 & 43.62 & 24.49 \\
    factGSG+mask & \textbf{24.23} & \textbf{7.69} & \textbf{38.88} & \textbf{35.14} \\
    \bottomrule
    \end{tabular}
    \vspace{-3pt}
    \caption{Zero-shot results when applying the connector to our pre-trained model (factGSG+mask) and PEGASUS*(GSG+mask). FactGSG+mask outperforms GSG+mask on all metrics.}
    \label{tab:zeroshot_factuality}
    \vspace{-5pt}
\end{table}
\begin{figure*}[ht]
    \centering
    \begin{subfigure}[]{0.32\textwidth}
    \centering
    \includegraphics[width=\textwidth]{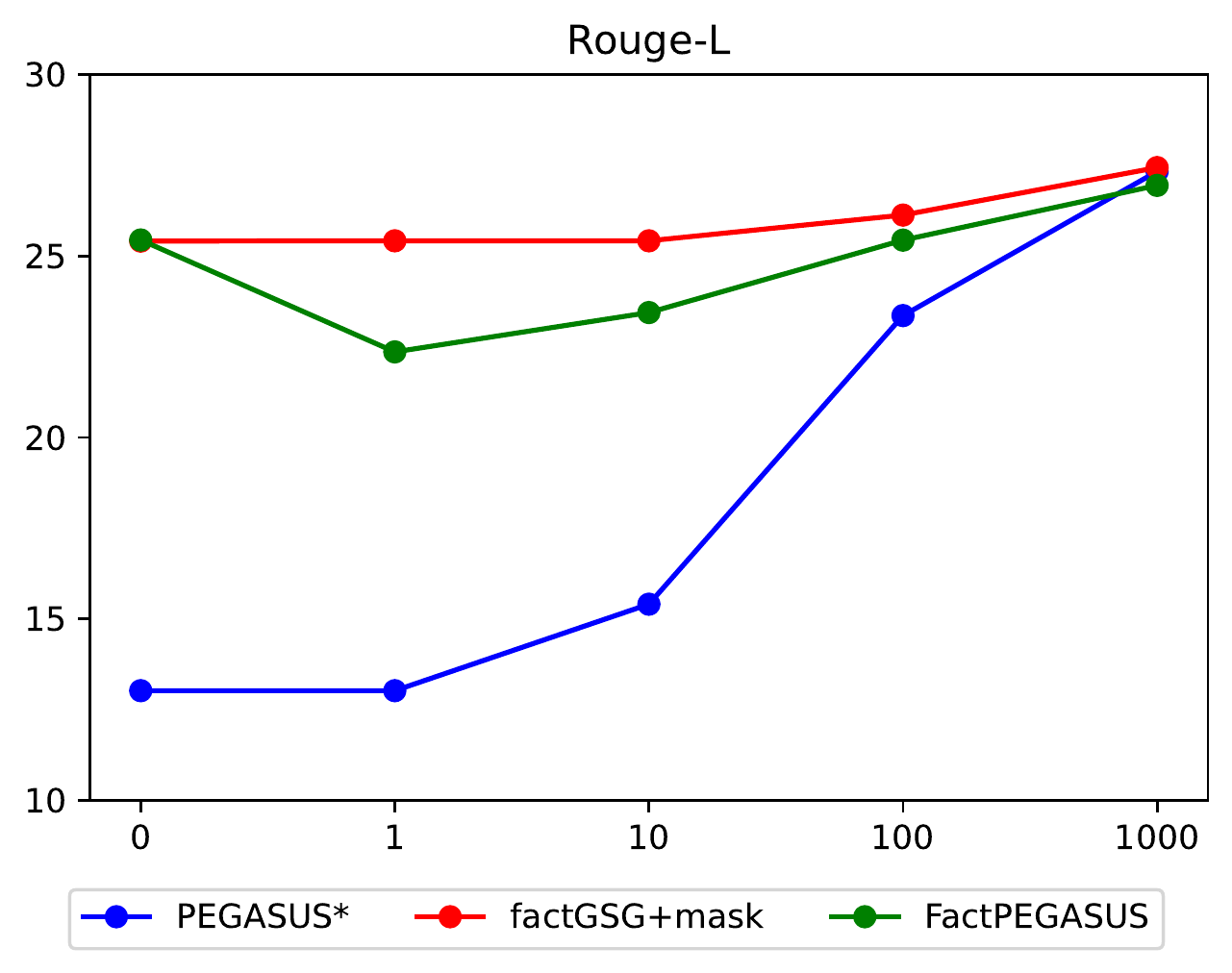}
    \end{subfigure}
    \begin{subfigure}[]{0.32\textwidth}
    \centering
    \includegraphics[width=\textwidth]{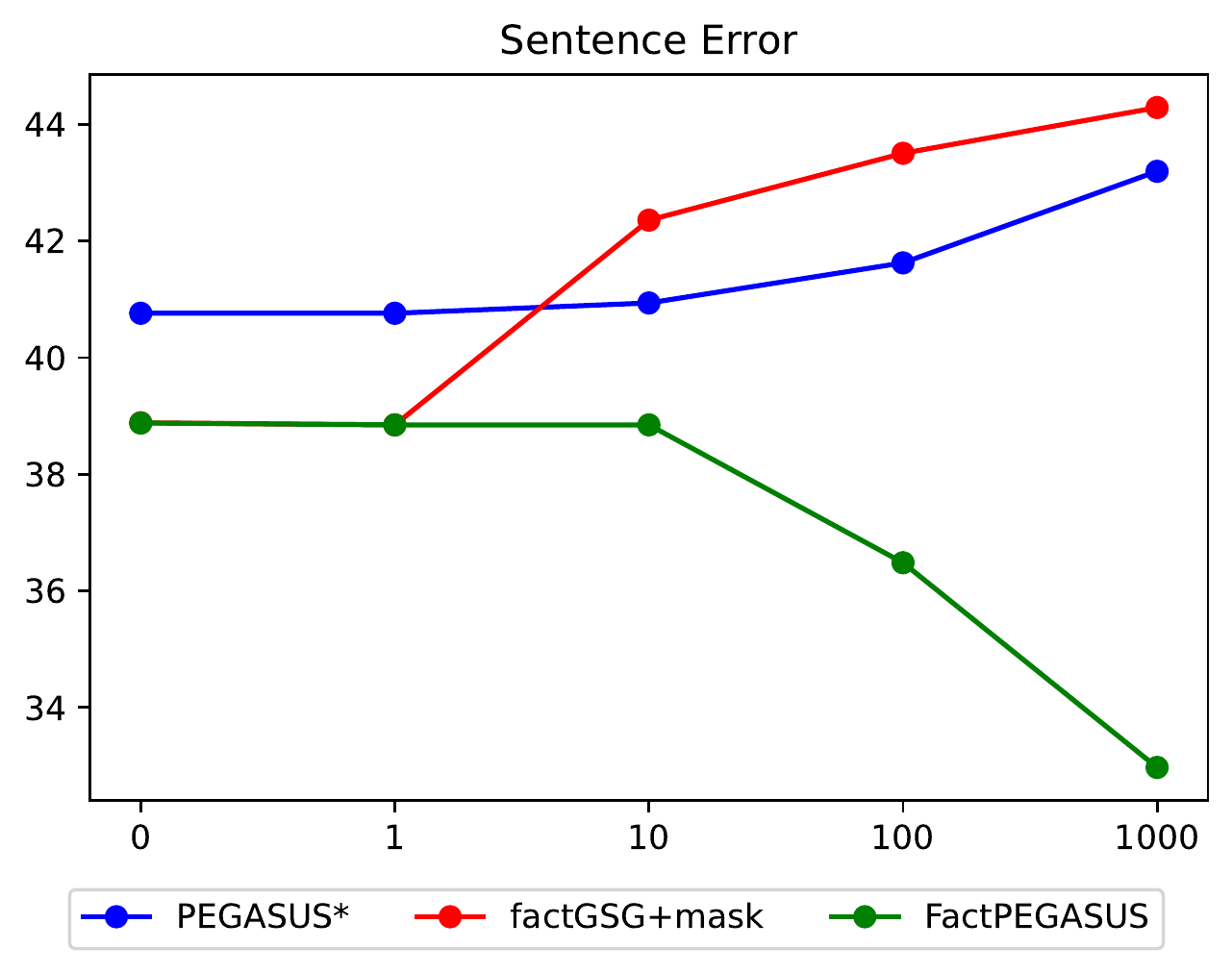}
    \end{subfigure}
    \begin{subfigure}[]{0.32\textwidth}
    \centering
    \includegraphics[width=\textwidth]{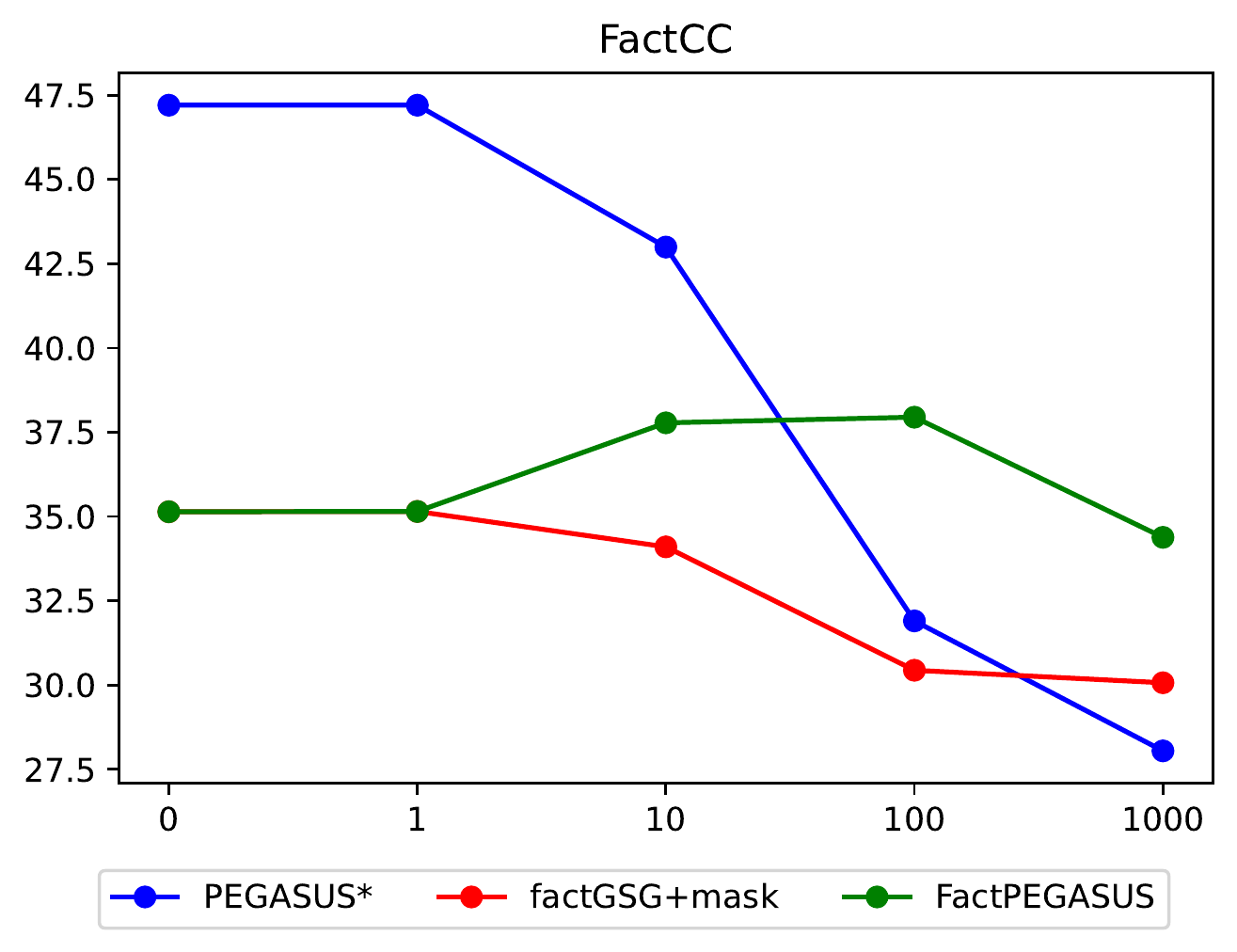}
    \end{subfigure}
    \vspace{-3pt}
    \caption{Zero-shot and few-shot results. The lines represent each models's performance when fine-tuned on 0 (zero-shot), 1, 10, 100, and 1000 examples.
    \MODEL{} consistently improves sentence error with more training data.
    Without the corrector and contrastor, factuality decreases with just 10 examples. }
    \label{fig:fewshot}
    \vspace{-5pt}
\end{figure*}

\subsection{Fine-tuning Ablation Studies}
\label{sec:ablation}

We present ablation studies of our proposed methods in \autoref{tab:ablation}. We first compare the performance of different strategies for the corrector and contrastor. For corrector, the level of aggressiveness in correcting hallucinations has a positive relationship with factuality metrics but a negative relationship with ROUGE-L. Although the remove method achieves the highest FactCC score, the combined method further lowers the token and sentence error while achieving relatively high ROUGE-L and FactCC. For contrastor, simulating intrinsic errors, which creates more challenging negative samples, provides better factuality results than simulating extrinsic ones. Finally, we show the additive gain in combining the best corrector and contrastor, as well as adding the connector to form the final model.

We report the same ablation studies for Gigaword and Wikihow in Appendix \ref{sec:ablation_gw_wh}, and that for PEGASUS* in Appendix \ref{sec:ablation_pegasus_star}.

\begin{figure*}[ht]
    \centering
    \small
    \begin{subfigure}[]{0.32\textwidth}
    \centering
    \includegraphics[width=\textwidth]{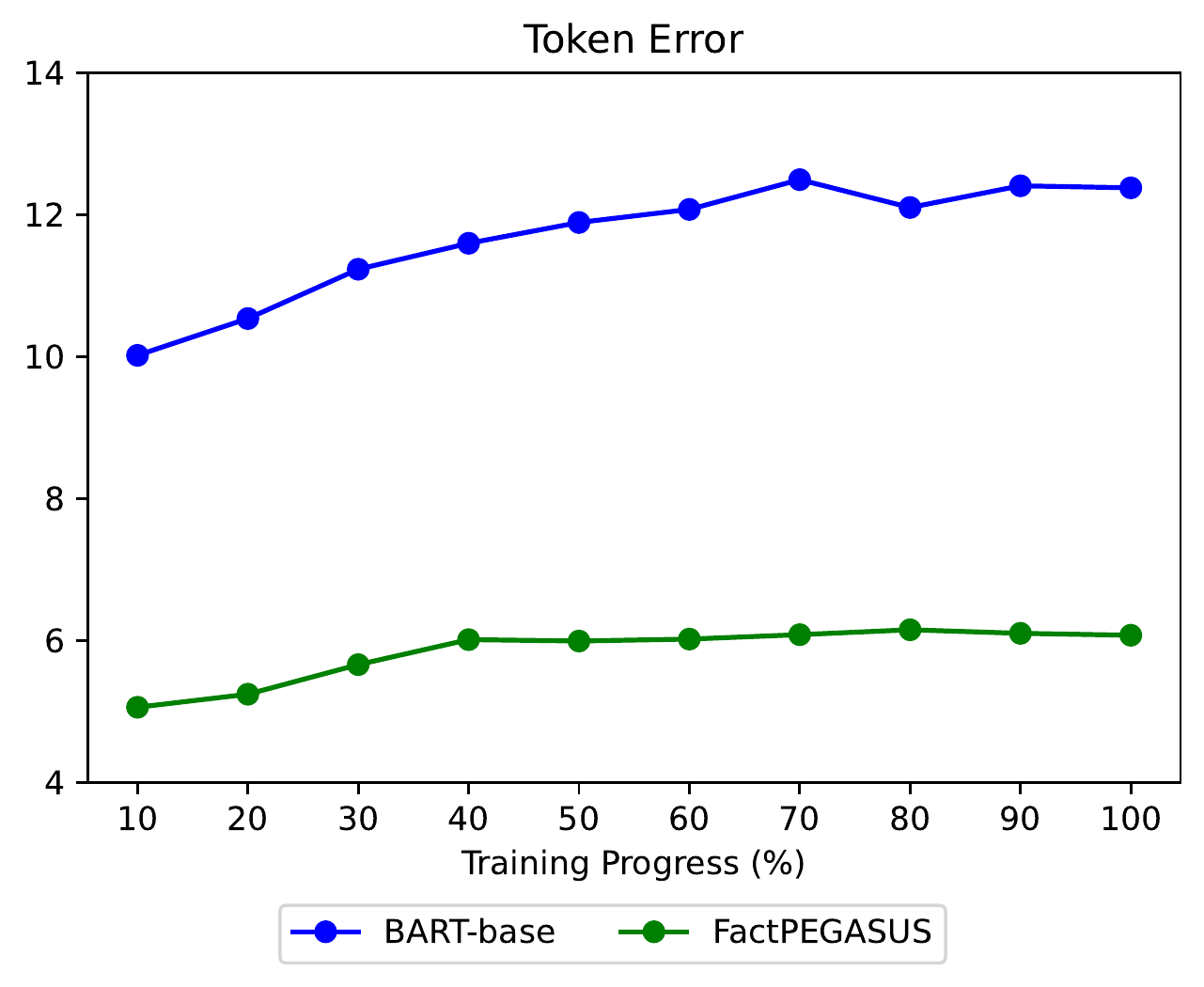}
    \end{subfigure}
    \begin{subfigure}[]{0.32\textwidth}
    \centering
    \includegraphics[width=\textwidth]{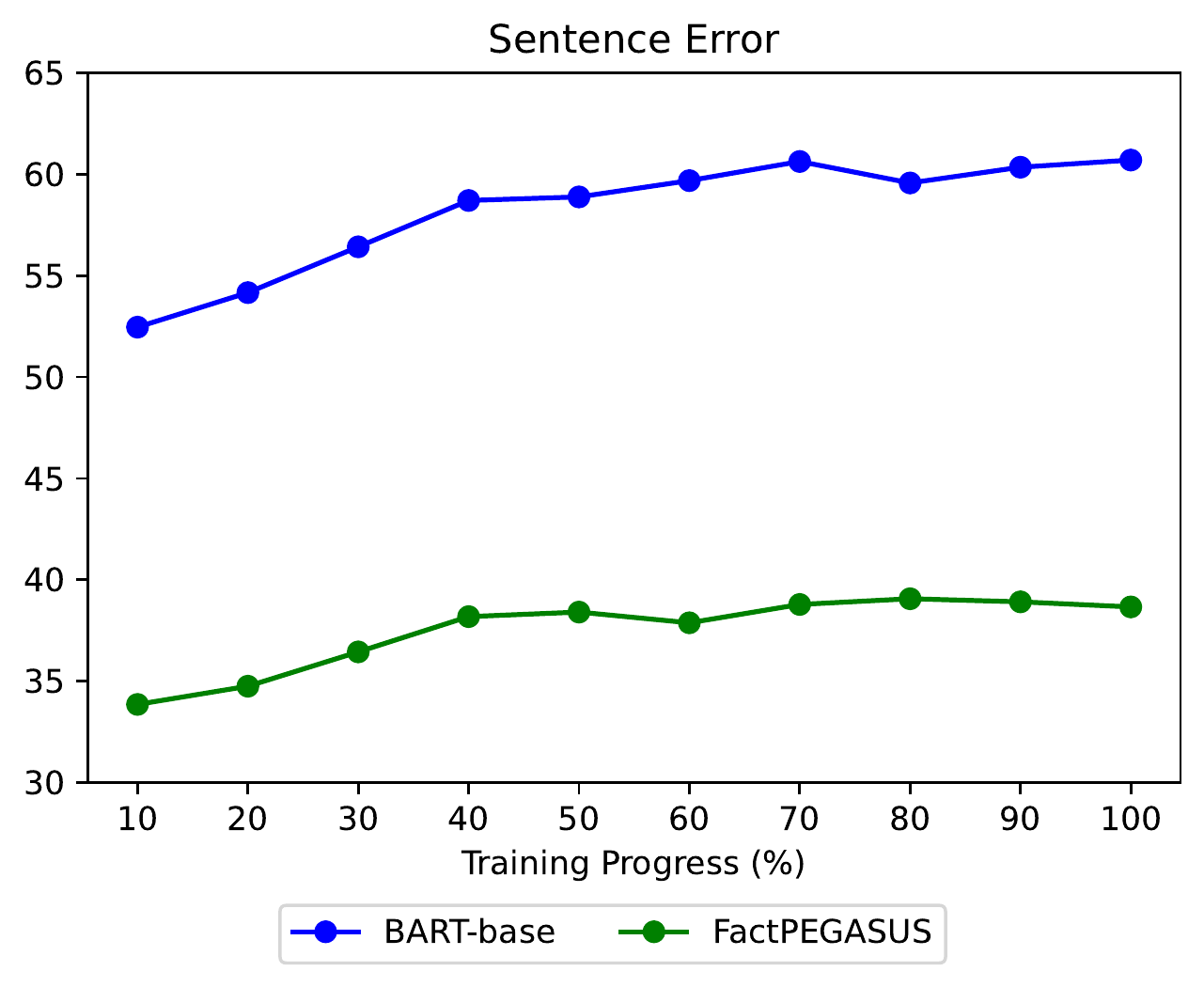}
    \end{subfigure}
    \begin{subfigure}[]{0.32\textwidth}
    \centering
    \includegraphics[width=\textwidth]{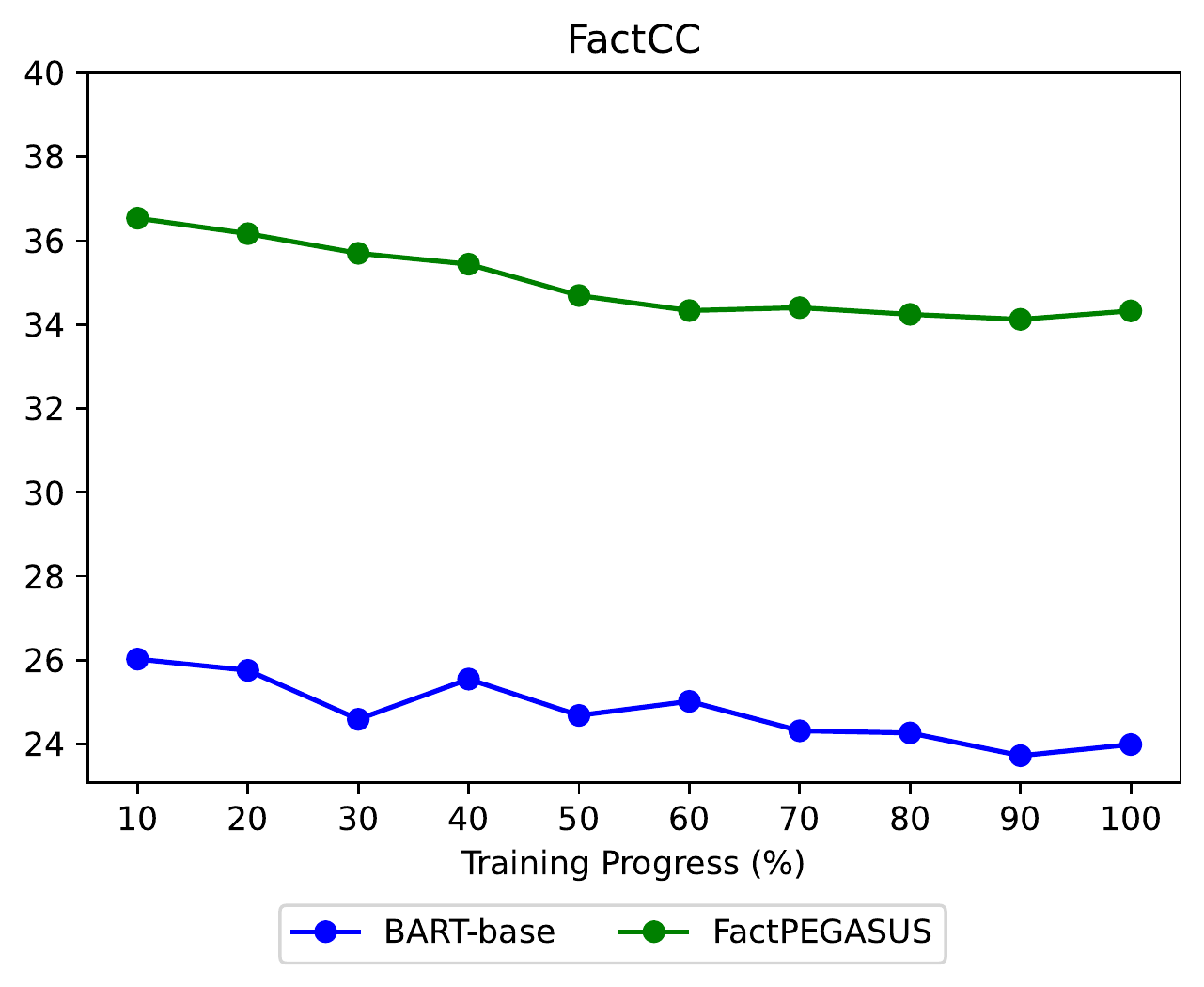}
    \end{subfigure}
    \vspace{-3pt}
    \caption{Factuality dynamics result. We show token error, sentence error, and FactCC as training progresses. \MODEL{} slows down factuality degradation for all metrics compared to BART-base.}
    \label{fig:training_progress}
    \vspace{-6pt}
\end{figure*}

\subsection{Zero-shot and Few-shot Results}
\label{sec:zeroshot_fewshot}

With the help of connector proposed in Section~\ref{sec:mask_finetuning}, we can explore how knowledge about factuality is transferred to fine-tuning, especially in the zero-shot and few-shot settings\footnote{Strictly speaking, typical zero-shot and few-shot settings do not allow using the full validation set. However, we use validation results to decide the position of the mask token.}.

\paragraph{Zero-Shot.} We apply the mask token to the best position and directly analyze the performance of the models on the test set. To better understand the effectiveness in transferring knowledge about summarization and factuality from the pre-training objective, we apply the connector to our pre-trained model (factGSG+mask) and PEGASUS* (GSG+mask), so that the two models differ only in their pre-training objective. We report the result in \autoref{tab:zeroshot_factuality}. FactGSG+mask outperforms GSG+mask on all metrics, especially for factuality metrics. Specifically, factGSG+mask lowers the sentence error by 5 points and increases FactCC by about 10 points. This observation confirms that the factGSG objective is more effective at capturing factuality than the original GSG objective.

\paragraph{Few-Shot.} We follow a similar setup in \citet{zhang2019pegasus}, where we limit the number of training data to 1, 10, 100, and 1,000, and then fine-tune the model up to 2,000 steps with the patience of 10 epochs for early stopping. We select the checkpoint with the best validation performance.

We conduct this experiment by comparing \MODEL{} to PEGASUS*, which has been shown for its ability to transfer with as little as 100 training examples \cite{zhang2019pegasus}. In addition, we report the performance of factGSG+mask to understand how the the model is affected without explicitly ensuring factuality (i.e. without corrector and contrastor). As shown in \autoref{fig:fewshot}, connector allows the model to better make use of the knowledge of pre-training and produces high-quality summaries, as both \MODEL{} and factGSG+mask produces a ROUGE-L score comparable to PEGASUS* trained with 1000 examples.

In terms of factuality, we notice that
with just 10 examples, PEGASUS* starts to degrade in factuality, which also applies to the factGSG+mask model. However, \MODEL{} demonstrates an opposite trajectory: Sentence error decreases with more training data, and FactCC remains about the same score.
This indicates that factual behavior is prone to be overwritten when factuality is not ensured explicitly, and thus calls for the importance of the corrector and contrastor.

\subsection{Factuality Dynamics during Fine-tuning}

To see whether the factuality degradation observed in few-shot experiment also applies to the full fine-tuning process,
we extend our analysis by studying the factuality dynamics, similar to \citet{goyal2021training}.
The authors observe an increase in sentence errors with the BART model during fine-tuning, and we analyze whether similar factuality degradation occurs for \MODEL{}. We save checkpoints of our models every $10\%$ of the total training steps, and evaluate the models on all three factuality metrics.
\autoref{fig:training_progress} shows the factuality dynamics during fine-tuning. We notice that the degradation occurs for both models but at a different degree. The token and sentence error for BART-base increase by 2 and 8 points, respectively. However, factuality for \MODEL{} remains similar, with only an increase of 1 point for token error and 4.8 points for sentence error. The degradation is only about half of what is observed with BART-base, indicating that \MODEL{} is better at avoiding learning nonfactual behaviors.

\begin{figure}[t]
    \centering
    \includegraphics[width=0.95\columnwidth,height=150pt, keepaspectratio]{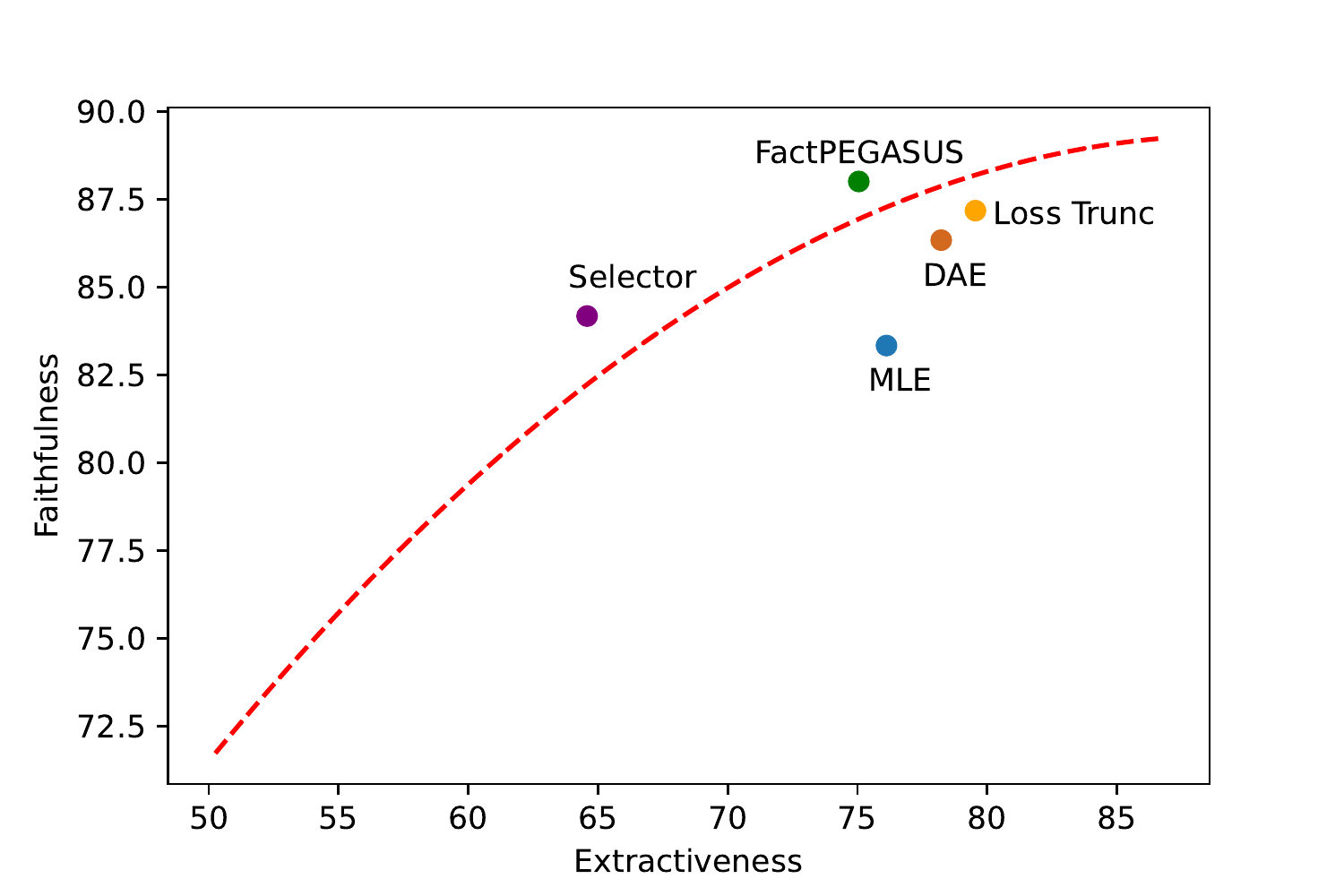}
    \vspace{-10pt}
    \caption{Faithfulness-abstractiveness trade-off curve, shown as the dashed red line, on Gigaword dataset.
    We plot each model's average faithfulness score evaluated by AMT against its extractiveness level.
    Our model lies above the graph, performing better than MLE-baseline, DAE \cite{goyal-durrett-2021-annotating}, and Loss Truncation \cite{kang-hashimoto-2020-improved}.
    }
    \label{fig:roc}
    \vspace{-6pt}
\end{figure}
\subsection{Factuality vs Abstractiveness Tradeoff}
Lastly, we wish to understand whether our proposed method is effectively improving factuality without relying on the increase in extractiveness. To this end, \citet{ladhak2021faithful} introduces a faithfulness-abstractiveness trade-off curve to measure the faithfulness given the model's extractiveness.
The authors kindly provided the same set of examples for Gigaword and AMT template for calculating the faithfulness score.

We show our result on Gigaword in \autoref{fig:roc}. We include the result of their proposed Selector and previous works, including Loss Truncation \cite{kang-hashimoto-2020-improved} and DAE \cite{goyal-durrett-2021-annotating}. We note that the baseline models increase factuality but mostly due to an increase in extractiveness and thus fall below the curve. In contrast, \MODEL{} lies above the line, indicating that we are effectively increasing factuality without relying much on becoming more extractive.

\section{Conclusion}
In this work, we proposed \MODEL{}, a model for abstractive summarization consisting of factuality-aware pre-training and modules for ensuring factuality during fine-tuning. We demonstrated the effectiveness of our model at improving factuality on three downstream abstractive summarization datasets, confirmed by our human evaluation. Our analysis showed that our proposed factuality-aware pre-training objective is effective at capturing knowledge of factuality compared to the original objective and that our fine-tuning modules reduce the factuality degradation observed with current models. We finally showed that improvement in factuality is not solely explained by the increase of extractiveness.

\section{Ethical Impact}
Our work aims at reducing the risk of generating hallucinations, and even possibly misinformation, for abstractive summarization models so that such models can be used safely for real-world applications. While we demonstrate that we can alleviate this problem, we stress that there is still a long way to go for improving factuality. Thus, we stress that such models should be used with caution for real-world applications.

\section*{Acknowledgment}

We thank the reviewers for their helpful comments. We also thank Shiyue Zhang and Xiang Zhou for useful discussions and comments on the paper. This  work  was  supported by NSF-CAREER Award 1846185 and NSF-AI Engage Institute DRL-211263.

\bibliography{anthology,custom}

\begin{thebibliography}{37}
\expandafter\ifx\csname natexlab\endcsname\relax\def\natexlab#1{#1}\fi

\bibitem[{Cao et~al.(2020)Cao, Dong, Wu, and Cheung}]{cao-etal-2020-factual}
Meng Cao, Yue Dong, Jiapeng Wu, and Jackie Chi~Kit Cheung. 2020.
\newblock \href {https://doi.org/10.18653/v1/2020.emnlp-main.506} {Factual
  error correction for abstractive summarization models}.
\newblock In \emph{Proceedings of the 2020 Conference on Empirical Methods in
  Natural Language Processing (EMNLP)}, pages 6251--6258, Online. Association
  for Computational Linguistics.

\bibitem[{Cao and Wang(2021)}]{cao-wang-2021-cliff}
Shuyang Cao and Lu~Wang. 2021.
\newblock \href {https://aclanthology.org/2021.emnlp-main.532} {{CLIFF}:
  Contrastive learning for improving faithfulness and factuality in abstractive
  summarization}.
\newblock In \emph{Proceedings of the 2021 Conference on Empirical Methods in
  Natural Language Processing}, pages 6633--6649, Online and Punta Cana,
  Dominican Republic. Association for Computational Linguistics.

\bibitem[{Chen et~al.(2021)Chen, Zhang, Sone, and
  Roth}]{chen-etal-2021-improving}
Sihao Chen, Fan Zhang, Kazoo Sone, and Dan Roth. 2021.
\newblock \href {https://doi.org/10.18653/v1/2021.naacl-main.475} {Improving
  faithfulness in abstractive summarization with contrast candidate generation
  and selection}.
\newblock In \emph{Proceedings of the 2021 Conference of the North American
  Chapter of the Association for Computational Linguistics: Human Language
  Technologies}, pages 5935--5941, Online. Association for Computational
  Linguistics.

\bibitem[{Chen et~al.(2020)Chen, Kornblith, Norouzi, and
  Hinton}]{chen2020simple}
Ting Chen, Simon Kornblith, Mohammad Norouzi, and Geoffrey Hinton. 2020.
\newblock \href {https://proceedings.mlr.press/v119/chen20j.html} {A simple
  framework for contrastive learning of visual representations}.
\newblock In \emph{Proceedings of the 37th International Conference on Machine
  Learning}, volume 119 of \emph{Proceedings of Machine Learning Research},
  pages 1597--1607. PMLR.

\bibitem[{Chopra et~al.(2005)Chopra, Hadsell, and LeCun}]{1467314}
S.~Chopra, R.~Hadsell, and Y.~LeCun. 2005.
\newblock \href {https://doi.org/10.1109/CVPR.2005.202} {Learning a similarity
  metric discriminatively, with application to face verification}.
\newblock In \emph{2005 IEEE Computer Society Conference on Computer Vision and
  Pattern Recognition (CVPR'05)}, volume~1, pages 539--546 vol. 1.

\bibitem[{Devlin et~al.(2019)Devlin, Chang, Lee, and
  Toutanova}]{devlin-etal-2019-bert}
Jacob Devlin, Ming-Wei Chang, Kenton Lee, and Kristina Toutanova. 2019.
\newblock \href {https://doi.org/10.18653/v1/N19-1423} {{BERT}: Pre-training of
  deep bidirectional transformers for language understanding}.
\newblock In \emph{Proceedings of the 2019 Conference of the North {A}merican
  Chapter of the Association for Computational Linguistics: Human Language
  Technologies, Volume 1 (Long and Short Papers)}, pages 4171--4186,
  Minneapolis, Minnesota. Association for Computational Linguistics.

\bibitem[{Dong et~al.(2020)Dong, Wang, Gan, Cheng, Cheung, and
  Liu}]{dong-etal-2020-multi}
Yue Dong, Shuohang Wang, Zhe Gan, Yu~Cheng, Jackie Chi~Kit Cheung, and Jingjing
  Liu. 2020.
\newblock \href {https://doi.org/10.18653/v1/2020.emnlp-main.749} {Multi-fact
  correction in abstractive text summarization}.
\newblock In \emph{Proceedings of the 2020 Conference on Empirical Methods in
  Natural Language Processing (EMNLP)}, pages 9320--9331, Online. Association
  for Computational Linguistics.

\bibitem[{Dreyer et~al.(2021)Dreyer, Liu, Nan, Atluri, and
  Ravi}]{dreyer2021analyzing}
Markus Dreyer, Mengwen Liu, Feng Nan, Sandeep Atluri, and Sujith Ravi. 2021.
\newblock \href {http://arxiv.org/abs/2108.02859} {Analyzing the
  abstractiveness-factuality tradeoff with nonlinear abstractiveness
  constraints}.
\newblock \emph{CoRR}, abs/2108.02859.

\bibitem[{Durmus et~al.(2020)Durmus, He, and Diab}]{durmus-etal-2020-feqa}
Esin Durmus, He~He, and Mona Diab. 2020.
\newblock \href {https://doi.org/10.18653/v1/2020.acl-main.454} {{FEQA}: A
  question answering evaluation framework for faithfulness assessment in
  abstractive summarization}.
\newblock In \emph{Proceedings of the 58th Annual Meeting of the Association
  for Computational Linguistics}, pages 5055--5070, Online. Association for
  Computational Linguistics.

\bibitem[{Efron and Tibshirani(1993)}]{EfroTibs93}
Bradley Efron and Robert~J. Tibshirani. 1993.
\newblock \emph{An Introduction to the Bootstrap}.
\newblock Number~57 in Monographs on Statistics and Applied Probability.
  Chapman \& Hall/CRC, Boca Raton, Florida, USA.

\bibitem[{Falke et~al.(2019)Falke, Ribeiro, Utama, Dagan, and
  Gurevych}]{falke-etal-2019-ranking}
Tobias Falke, Leonardo F.~R. Ribeiro, Prasetya~Ajie Utama, Ido Dagan, and Iryna
  Gurevych. 2019.
\newblock \href {https://doi.org/10.18653/v1/P19-1213} {Ranking generated
  summaries by correctness: An interesting but challenging application for
  natural language inference}.
\newblock In \emph{Proceedings of the 57th Annual Meeting of the Association
  for Computational Linguistics}, pages 2214--2220, Florence, Italy.
  Association for Computational Linguistics.

\bibitem[{Goyal and Durrett(2021)}]{goyal-durrett-2021-annotating}
Tanya Goyal and Greg Durrett. 2021.
\newblock \href {https://doi.org/10.18653/v1/2021.naacl-main.114} {Annotating
  and modeling fine-grained factuality in summarization}.
\newblock In \emph{Proceedings of the 2021 Conference of the North American
  Chapter of the Association for Computational Linguistics: Human Language
  Technologies}, pages 1449--1462, Online. Association for Computational
  Linguistics.

\bibitem[{Goyal et~al.(2022)Goyal, Xu, Li, and Durrett}]{goyal2021training}
Tanya Goyal, Jiacheng Xu, Junyi~Jessy Li, and Greg Durrett. 2022.
\newblock Training dynamics for text summarization models.
\newblock In \emph{Proceedings of ACL}.

\bibitem[{Kang and Hashimoto(2020)}]{kang-hashimoto-2020-improved}
Daniel Kang and Tatsunori~B. Hashimoto. 2020.
\newblock \href {https://doi.org/10.18653/v1/2020.acl-main.66} {Improved
  natural language generation via loss truncation}.
\newblock In \emph{Proceedings of the 58th Annual Meeting of the Association
  for Computational Linguistics}, pages 718--731, Online. Association for
  Computational Linguistics.

\bibitem[{Koupaee and Wang(2018)}]{koupaee2018wikihow}
Mahnaz Koupaee and William~Yang Wang. 2018.
\newblock \href {http://arxiv.org/abs/1810.09305} {Wikihow: A large scale text
  summarization dataset}.

\bibitem[{Kryscinski et~al.(2019)Kryscinski, Keskar, McCann, Xiong, and
  Socher}]{kryscinski-etal-2019-neural}
Wojciech Kryscinski, Nitish~Shirish Keskar, Bryan McCann, Caiming Xiong, and
  Richard Socher. 2019.
\newblock \href {https://doi.org/10.18653/v1/D19-1051} {Neural text
  summarization: A critical evaluation}.
\newblock In \emph{Proceedings of the 2019 Conference on Empirical Methods in
  Natural Language Processing and the 9th International Joint Conference on
  Natural Language Processing (EMNLP-IJCNLP)}, pages 540--551, Hong Kong,
  China. Association for Computational Linguistics.

\bibitem[{Kryscinski et~al.(2020)Kryscinski, McCann, Xiong, and
  Socher}]{kryscinski-etal-2020-evaluating}
Wojciech Kryscinski, Bryan McCann, Caiming Xiong, and Richard Socher. 2020.
\newblock \href {https://doi.org/10.18653/v1/2020.emnlp-main.750} {Evaluating
  the factual consistency of abstractive text summarization}.
\newblock In \emph{Proceedings of the 2020 Conference on Empirical Methods in
  Natural Language Processing (EMNLP)}, pages 9332--9346, Online. Association
  for Computational Linguistics.

\bibitem[{Kudo(2018)}]{kudo-2018-subword}
Taku Kudo. 2018.
\newblock \href {https://doi.org/10.18653/v1/P18-1007} {Subword regularization:
  Improving neural network translation models with multiple subword
  candidates}.
\newblock In \emph{Proceedings of the 56th Annual Meeting of the Association
  for Computational Linguistics (Volume 1: Long Papers)}, pages 66--75,
  Melbourne, Australia. Association for Computational Linguistics.

\bibitem[{Ladhak et~al.(2021)Ladhak, Durmus, He, Cardie, and
  McKeown}]{ladhak2021faithful}
Faisal Ladhak, Esin Durmus, He~He, Claire Cardie, and Kathleen McKeown. 2021.
\newblock \href {http://arxiv.org/abs/2108.13684} {Faithful or extractive? on
  mitigating the faithfulness-abstractiveness trade-off in abstractive
  summarization}.

\bibitem[{Lee et~al.(2021)Lee, Lee, and Hwang}]{lee2021contrastive}
Seanie Lee, Dong~Bok Lee, and Sung~Ju Hwang. 2021.
\newblock \href {https://openreview.net/forum?id=Wga_hrCa3P3} {Contrastive
  learning with adversarial perturbations for conditional text generation}.
\newblock In \emph{International Conference on Learning Representations}.

\bibitem[{Lewis et~al.(2020)Lewis, Liu, Goyal, Ghazvininejad, Mohamed, Levy,
  Stoyanov, and Zettlemoyer}]{lewis-etal-2020-bart}
Mike Lewis, Yinhan Liu, Naman Goyal, Marjan Ghazvininejad, Abdelrahman Mohamed,
  Omer Levy, Veselin Stoyanov, and Luke Zettlemoyer. 2020.
\newblock \href {https://doi.org/10.18653/v1/2020.acl-main.703} {{BART}:
  Denoising sequence-to-sequence pre-training for natural language generation,
  translation, and comprehension}.
\newblock In \emph{Proceedings of the 58th Annual Meeting of the Association
  for Computational Linguistics}, pages 7871--7880, Online. Association for
  Computational Linguistics.

\bibitem[{Lhoest et~al.(2021)Lhoest, Villanova~del Moral, Jernite, Thakur, von
  Platen, Patil, Chaumond, Drame, Plu, Tunstall, Davison, {\v{S}}a{\v{s}}ko,
  Chhablani, Malik, Brandeis, Le~Scao, Sanh, Xu, Patry, McMillan-Major, Schmid,
  Gugger, Delangue, Matussi{\`e}re, Debut, Bekman, Cistac, Goehringer, Mustar,
  Lagunas, Rush, and Wolf}]{lhoest-etal-2021-datasets}
Quentin Lhoest, Albert Villanova~del Moral, Yacine Jernite, Abhishek Thakur,
  Patrick von Platen, Suraj Patil, Julien Chaumond, Mariama Drame, Julien Plu,
  Lewis Tunstall, Joe Davison, Mario {\v{S}}a{\v{s}}ko, Gunjan Chhablani,
  Bhavitvya Malik, Simon Brandeis, Teven Le~Scao, Victor Sanh, Canwen Xu,
  Nicolas Patry, Angelina McMillan-Major, Philipp Schmid, Sylvain Gugger,
  Cl{\'e}ment Delangue, Th{\'e}o Matussi{\`e}re, Lysandre Debut, Stas Bekman,
  Pierric Cistac, Thibault Goehringer, Victor Mustar, Fran{\c{c}}ois Lagunas,
  Alexander Rush, and Thomas Wolf. 2021.
\newblock \href {http://arxiv.org/abs/2109.02846} {Datasets: A community
  library for natural language processing}.
\newblock In \emph{Proceedings of the 2021 Conference on Empirical Methods in
  Natural Language Processing: System Demonstrations}, pages 175--184, Online
  and Punta Cana, Dominican Republic. Association for Computational
  Linguistics.

\bibitem[{Lin(2004)}]{lin-2004-rouge}
Chin-Yew Lin. 2004.
\newblock \href {https://aclanthology.org/W04-1013} {{ROUGE}: A package for
  automatic evaluation of summaries}.
\newblock In \emph{Text Summarization Branches Out}, pages 74--81, Barcelona,
  Spain. Association for Computational Linguistics.

\bibitem[{Lin et~al.(2021)Lin, Hilton, and Evans}]{lin2021truthfulqa}
Stephanie Lin, Jacob Hilton, and Owain Evans. 2021.
\newblock \href {http://arxiv.org/abs/2109.07958} {Truthfulqa: Measuring how
  models mimic human falsehoods}.

\bibitem[{Liu and Liu(2021)}]{liu-liu-2021-simcls}
Yixin Liu and Pengfei Liu. 2021.
\newblock \href {https://doi.org/10.18653/v1/2021.acl-short.135} {{S}im{CLS}: A
  simple framework for contrastive learning of abstractive summarization}.
\newblock In \emph{Proceedings of the 59th Annual Meeting of the Association
  for Computational Linguistics and the 11th International Joint Conference on
  Natural Language Processing (Volume 2: Short Papers)}, pages 1065--1072,
  Online. Association for Computational Linguistics.

\bibitem[{Logan et~al.(2021)Logan, Balažević, Wallace, Petroni, Singh, and
  Riedel}]{logan2021cutting}
Robert~L. Logan, Ivana Balažević, Eric Wallace, Fabio Petroni, Sameer Singh,
  and Sebastian Riedel. 2021.
\newblock \href {http://arxiv.org/abs/2106.13353} {Cutting down on prompts and
  parameters: Simple few-shot learning with language models}.

\bibitem[{Maynez et~al.(2020)Maynez, Narayan, Bohnet, and
  McDonald}]{maynez-etal-2020-faithfulness}
Joshua Maynez, Shashi Narayan, Bernd Bohnet, and Ryan McDonald. 2020.
\newblock \href {https://doi.org/10.18653/v1/2020.acl-main.173} {On
  faithfulness and factuality in abstractive summarization}.
\newblock In \emph{Proceedings of the 58th Annual Meeting of the Association
  for Computational Linguistics}, pages 1906--1919, Online. Association for
  Computational Linguistics.

\bibitem[{Nan et~al.(2021)Nan, Nallapati, Wang, Nogueira~dos Santos, Zhu,
  Zhang, McKeown, and Xiang}]{nan-etal-2021-entity}
Feng Nan, Ramesh Nallapati, Zhiguo Wang, Cicero Nogueira~dos Santos, Henghui
  Zhu, Dejiao Zhang, Kathleen McKeown, and Bing Xiang. 2021.
\newblock \href {https://aclanthology.org/2021.eacl-main.235} {Entity-level
  factual consistency of abstractive text summarization}.
\newblock In \emph{Proceedings of the 16th Conference of the European Chapter
  of the Association for Computational Linguistics: Main Volume}, pages
  2727--2733, Online. Association for Computational Linguistics.

\bibitem[{Narayan et~al.(2018)Narayan, Cohen, and
  Lapata}]{narayan-etal-2018-dont}
Shashi Narayan, Shay~B. Cohen, and Mirella Lapata. 2018.
\newblock \href {https://doi.org/10.18653/v1/D18-1206} {Don{'}t give me the
  details, just the summary! topic-aware convolutional neural networks for
  extreme summarization}.
\newblock In \emph{Proceedings of the 2018 Conference on Empirical Methods in
  Natural Language Processing}, pages 1797--1807, Brussels, Belgium.
  Association for Computational Linguistics.

\bibitem[{Pagnoni et~al.(2021)Pagnoni, Balachandran, and
  Tsvetkov}]{pagnoni-etal-2021-understanding}
Artidoro Pagnoni, Vidhisha Balachandran, and Yulia Tsvetkov. 2021.
\newblock \href {https://doi.org/10.18653/v1/2021.naacl-main.383}
  {Understanding factuality in abstractive summarization with {FRANK}: A
  benchmark for factuality metrics}.
\newblock In \emph{Proceedings of the 2021 Conference of the North American
  Chapter of the Association for Computational Linguistics: Human Language
  Technologies}, pages 4812--4829, Online. Association for Computational
  Linguistics.

\bibitem[{Raffel et~al.(2020)Raffel, Shazeer, Roberts, Lee, Narang, Matena,
  Zhou, Li, and Liu}]{2020t5}
Colin Raffel, Noam Shazeer, Adam Roberts, Katherine Lee, Sharan Narang, Michael
  Matena, Yanqi Zhou, Wei Li, and Peter~J. Liu. 2020.
\newblock \href {http://jmlr.org/papers/v21/20-074.html} {Exploring the limits
  of transfer learning with a unified text-to-text transformer}.
\newblock \emph{Journal of Machine Learning Research}, 21(140):1--67.

\bibitem[{Reimers and Gurevych(2019)}]{reimers-gurevych-2019-sentence}
Nils Reimers and Iryna Gurevych. 2019.
\newblock \href {https://doi.org/10.18653/v1/D19-1410} {Sentence-{BERT}:
  Sentence embeddings using {S}iamese {BERT}-networks}.
\newblock In \emph{Proceedings of the 2019 Conference on Empirical Methods in
  Natural Language Processing and the 9th International Joint Conference on
  Natural Language Processing (EMNLP-IJCNLP)}, pages 3982--3992, Hong Kong,
  China. Association for Computational Linguistics.

\bibitem[{Rush et~al.(2015)Rush, Chopra, and Weston}]{rush-etal-2015-neural}
Alexander~M. Rush, Sumit Chopra, and Jason Weston. 2015.
\newblock \href {https://doi.org/10.18653/v1/D15-1044} {A neural attention
  model for abstractive sentence summarization}.
\newblock In \emph{Proceedings of the 2015 Conference on Empirical Methods in
  Natural Language Processing}, pages 379--389, Lisbon, Portugal. Association
  for Computational Linguistics.

\bibitem[{Scialom et~al.(2021)Scialom, Dray, Lamprier, Piwowarski, Staiano,
  Wang, and Gallinari}]{scialom-etal-2021-questeval}
Thomas Scialom, Paul-Alexis Dray, Sylvain Lamprier, Benjamin Piwowarski, Jacopo
  Staiano, Alex Wang, and Patrick Gallinari. 2021.
\newblock \href {https://aclanthology.org/2021.emnlp-main.529} {{Q}uest{E}val:
  Summarization asks for fact-based evaluation}.
\newblock In \emph{Proceedings of the 2021 Conference on Empirical Methods in
  Natural Language Processing}, pages 6594--6604, Online and Punta Cana,
  Dominican Republic. Association for Computational Linguistics.

\bibitem[{Wang et~al.(2020)Wang, Cho, and Lewis}]{wang-etal-2020-asking}
Alex Wang, Kyunghyun Cho, and Mike Lewis. 2020.
\newblock \href {https://doi.org/10.18653/v1/2020.acl-main.450} {Asking and
  answering questions to evaluate the factual consistency of summaries}.
\newblock In \emph{Proceedings of the 58th Annual Meeting of the Association
  for Computational Linguistics}, pages 5008--5020, Online. Association for
  Computational Linguistics.

\bibitem[{Wolf et~al.(2020)Wolf, Debut, Sanh, Chaumond, Delangue, Moi, Cistac,
  Rault, Louf, Funtowicz, Davison, Shleifer, von Platen, Ma, Jernite, Plu, Xu,
  Le~Scao, Gugger, Drame, Lhoest, and Rush}]{wolf-etal-2020-transformers}
Thomas Wolf, Lysandre Debut, Victor Sanh, Julien Chaumond, Clement Delangue,
  Anthony Moi, Pierric Cistac, Tim Rault, Remi Louf, Morgan Funtowicz, Joe
  Davison, Sam Shleifer, Patrick von Platen, Clara Ma, Yacine Jernite, Julien
  Plu, Canwen Xu, Teven Le~Scao, Sylvain Gugger, Mariama Drame, Quentin Lhoest,
  and Alexander Rush. 2020.
\newblock \href {https://doi.org/10.18653/v1/2020.emnlp-demos.6} {Transformers:
  State-of-the-art natural language processing}.
\newblock In \emph{Proceedings of the 2020 Conference on Empirical Methods in
  Natural Language Processing: System Demonstrations}, pages 38--45, Online.
  Association for Computational Linguistics.

\bibitem[{Zhang et~al.(2020)Zhang, Zhao, Saleh, and Liu}]{zhang2019pegasus}
Jingqing Zhang, Yao Zhao, Mohammad Saleh, and Peter~J. Liu. 2020.
\newblock Pegasus: Pre-training with extracted gap-sentences for abstractive
  summarization.
\newblock In \emph{Proceedings of the 37th International Conference on Machine
  Learning}, ICML'20. JMLR.org.

\end{thebibliography}
\bibliographystyle{acl_natbib}

\appendix

\section{More Details on Experimental Setup}\label{sec:experiment_details}

\begin{table*}[t]
    \centering
    \small
    \begin{tabular}{cccccc}
    \toprule
      Dataset & Train & Validation & Test & \multicolumn{2}{c}{Corrector}  \\
       & & & & Replace & Remove \\
     \midrule
     C4 & 364,868,892 & 364,608 & - & - & - \\
     \textit{realnewslike} & 13,799,838 & 13,863 & - & - & - \\
     \midrule
     XSum & 204,045 & 11,332 & 11,334 & 54,036 & 152,716 \\
     WikiHow & 157,252 & 5,559 & 5,577 & 8,077 & 71,936 \\
     Gigaword & 3,803,957 & 178,651 & 1,951 & 115,896 & 1,296,168  \\
     \bottomrule
    \end{tabular}
    \caption{Dataset Statistics. We show the number of exmaples in each split, as well as the number of training examples changed using the replace and remove strategy of the  corrector.}
    \label{tab:dataset_stat}
\end{table*}
\subsection{Datasets}\label{sec:dataset_details}
Following PEGASUS, we pre-train on the C4 dataset, a large collection of documents from Common Crawl. We evaluate our pre-trained model on three downstream abstractive summarization datasets: XSum, WikiHow, and Gigaword. XSum is a collection of articles from the British Broadcasting Corporation, Gigaword is a large collection of news articles headlines, and WikiHow consists of how-to articles.

We show the dataset statistics for pre-training and fine-tuning in \autoref{tab:dataset_stat}, where we present the number of examples in the training, validation, and test splits. We also show the number of examples corrected using the replace and remove method. All datasets are from \textit{datasets} \cite{lhoest-etal-2021-datasets}.

\subsection{Evaluation Metrics}

We use the ROUGE package provided by \textit{transformers} \cite{wolf-etal-2020-transformers}. We follow the instructions provided by the authors of the factuality metrics to set up and run their code. We report all scores of our models from single runs.

\subsection{Training Details}
We use \textit{transformers} library for the training script and the checkpoints of the pre-trained models. We use the default setting, including the AdamW optimizer and the linear rate scheduler. We also use mixed precision for both pre-training and fine-tuning the models. We conduct our experiments on the RTX A6000 GPU with 48GB memory and the A100 GPU with 40GB memory. BART-base model has 139M parameters, and PEGASUS* and \MODEL{} have 175M parameters.

\subsection{Pre-training Setup}\label{sec:pretraining_detail}
\paragraph{Model Architecture.} We use the same architecture as BART-base. Specifically, the model has $L=6$, $H=768$, $F=3072$, $A=12$, where $L$ is the number of layers, $H$ is the hidden size, $F$ is the dimension for feed-forward layer, and $A$ is the number of self-attention heads. We use the SentencePiece \cite{kudo-2018-subword} unigram model tokenizer from PEGASUS with a vocabulary size of 96,103. 

\paragraph{Sentence Selection Criteria.} Before pre-training the full model, we first determine the best sentence selection criteria that produces more factual summaries with comparable quality.
We experiment with sentence selection criteria that use ROUGE-1, ROUGE-2, and ROUGE-L, as well as combining each with FactCC.
To understand the effect of the pre-training objective on factuality directly, we evaluate the performance on the XSum dataset without applying any of our proposed fine-tuning modules. Following \citet{zhang2019pegasus}, we report the models' relative performance to the base model, which only uses ROUGE-1 as the selection criteria. We use the normalized ROUGE F1 scores $\frac{1}{3} (\frac{R1}{R1_{base}} + \frac{R2}{R2_{base}} + \frac{RL}{RL_{base}})$, where $R1_{base}$, $R2_{base}$, and $RL_{base}$ are the ROUGE F1 scores of the base model.
We similarly report the factuality metrics by normalizing each score by that of the base model. We take the complement of token error and sentence error as token accuracy and sentence accuracy, respectively, to present all metrics where higher is better. 

Similar to previous works \cite{lewis-etal-2020-bart, zhang2019pegasus, 2020t5} that save computational resources when selecting strategies for pre-training, we pre-train these model on the \textit{realnewslike} subset of the C4 dataset with less steps.

\paragraph{Pre-training Details.} We use a learning rate of 1e-4, a weight decay of 0.01, and set the maximum number of input tokens to be 512 and a maximum number of output tokens to be 256. We use a batch size of 256. We pre-train the full model for 750,000 steps with a warm-up of 20,000 steps, and only pre-train the smaller models for the sentence selection criteria experiment for 250,000 steps. Pre-training the smaller models takes 30 hours, and pre-training the full model takes 90 hours. 

\paragraph{Calculating FactCC Score.} In practice, running FactCC on each sentence-document pair of the pre-training data is expensive. Thus, we opt to only calculate
the FactCC score for the top 5 sentences according to the ROUGE score between the sentence and the rest of the document.

\begin{table*}[t]
    \centering
    \resizebox{\textwidth}{!}{
    \begin{tabular}{c c c c c c c}
    \toprule
    Dataset & Learning rate & Num Steps & Warmup & Batch size & Max Input tokens & Max Target tokens \\
    \midrule 
    XSum  & 3e-5 & 15k & 500 & 256 & 512 & 64  \\
    WikiHow & 3e-5 & 15k & 500 & 256 & 512 & 256 \\
    Gigaword & 3e-5 & 50k & 2000 & 256 & 128 & 32\\
    \bottomrule
    \end{tabular}
    }
    \caption{Hyperparametrs for fine-tuning on the three tasks.}
    \label{tab:hyperparam}
\end{table*}
\subsection{Fine-tuning Setup}\label{sec:finetuning_detail}

For all datasets, we use a label smoothing of 0.1. For decoding, we use a beam size of 6 for all datasets. Task-specific hyper-parameters are shown in \autoref{tab:hyperparam}. Fine-tuning on XSum and WikiHow takes 8 hours, and fine-tuning on Gigaword takes 11 hours. Decoding on XSum and Gigaword takes half an hour, while decoding WikiHow takes an hour. We use 5 negative examples for the contrastor and set $\lambda$ to 5 when calculating the combined loss. We set the temperature $\tau$ to 0.05.

For fine-tuning DAE and CLIFF, we follow the authors' instructions and fine-tune BART-base with their respective code and hyper-parameters. For WikiHow and Gigaword, we use the same hyperparameters as above.

\section{Implementation Details for Corrector and Contrastor}
\subsection{Corrector}\label{sec:corrector_details}
We use spaCy's NER model\footnote{We use the en\_core\_web\_trf model.} to find entities in the document and summary. Entities in the summary sentence are considered nonfactual if no matching document entities with the same string are found. We have previously experimented with the additional requirement of matching entity type similar to \citet{kryscinski-etal-2020-evaluating}, but we find that this constraint unintentionally causes some correct entities to be considered hallucinating, leading to unnecessarily less informative summaries when removed.

Given hallucinated entities, we can perform either replace or remove operations. For replace, we find document entities whose words are all contained in the selected entity.

For the remove method, we need to make sure to also remove any related words. We use spaCy's dependency parser to systematically remove those. The algorithm is as follows: We first add all the tokens in the selected hallucinated entity to the list of tokens to remove. Then, we recursively find all parents that contain the dependency relation of \textit{pobj} and \textit{prep} without any other children and add those to the tokens to remove. Finally, we add all children that do not have the label \textit{compound}, \textit{relcl}, and \textit{fixed}. The final set of words will then be removed in the summary sentence.

We qualitatively observe that this approach can cover most of the edge cases that would otherwise result in ungrammatical sentences. Nevertheless, this method is not perfect.
We include some sample output with the remove method in \autoref{fig:corrector_example}. The algorithm is good at removing entities and related words, such as prepositions, as illustrated in example 1, 3, and 5. However, we observe that it will create ungrammatical sentences when the hallucinated entity is the subject (example 2), or the object of a transitive verb (example 6).

We leave exploration with the best systematic correction algorithm or models for future work.

\subsection{Contrastor}\label{sec:contrastor_details}
Similar to \citet{kryscinski-etal-2020-evaluating}, we generate hallucinated summaries by performing entity perturbation on the original summaries. We find entity candidates using the NER labels and sort them into three categories: We include MONEY, QUANTITY, and CARDINAL as \textit{number}, DATE and TIME as \textit{date}, and all other labels as \textit{named entities}. We randomly select a factual entity in the summary and replace it with an entity belonging to the same category.

For extrinsic hallucinations, we sample candidates of the same category from the training corpus but exclude those present in the document. For the intrinsic case, we select to consider the entities from the document. The number of negative examples for all tasks is 5.

\begin{table*}[t]
    \centering
    \small
    \begin{tabular}{c | c c c  | c c c  | c c c  }
    \toprule
    & \multicolumn{3}{c}{XSum} &  \multicolumn{3}{c}{WikiHow} & \multicolumn{3}{c}{Gigaword}  \\
    Pos. & R1 & R2 & RL & R1 & R2 & RL & R1 & R2 & RL \\
    \midrule
    1 & \textbf{32.84} & \textbf{11.32} & \textbf{25.35} & \textbf{21.02} & \textbf{4.85} & \textbf{4.85} & \textbf{26.19} & \textbf{9.09} & \textbf{22.92} \\
    2 & 24.10 & 5.90 & 18.02 & 20.65 & 4.80 & 14.80 & 22.89 & 7.22 & 20.03 \\
    3 & 21.23 & 4.30 & 15.69 & 20.81 & 4.89 & 14.93 & 22.89 & 7.22 & 20.03 \\
    4 & 19.52 & 3.47 & 14.41 & 20.61 & 4.79 & 14.77 & 22.89	& 7.22 & 20.03 \\
    5 & 18.77 & 3.03 & 13.86 & 20.72 & 4.85 & 14.82 & 22.89	& 7.22 & 20.03 \\
    6 & 18.22 & 2.80 & 13.51 & 20.69 & 4.82 & 14.87 & 22.89	& 7.22 & 20.03 \\
    \bottomrule
    \end{tabular}
    \caption{ROUGE score on validation set when the mask token is placed at different position. Pos. indicates placing the mask token before the $i$th sentence. Pos. 1 indicates the beginning of the document.}
    \label{tab:mask_token_position}
\end{table*}
\section{Connector Result}
\label{sec:mask_token_position}
This mask-token fine-tuning technique can be seen as a form of prompting, where we elicit our desired faithful abstractive summarization behavior from the pre-trained model directly.
Specifically, we consider this as null-prompting \cite{logan2021cutting}, where using the mask token as the prompt can achieve competitive results with manually engineered prompts. Conveniently, since the mask token during pre-training already serves as a placeholder of where the summary sentence should be generated, it naturally serves as a valid prompt. \autoref{fig:model_finetuning} shows an example of adding the mask token before the first sentence and thus creating a similar setup for pre-training.

\begin{table}[t]
    \centering
    \small
    \begin{tabular}{l c c c c}
    \toprule
     Model & R1 & R2 & RL \\
    \midrule
    BART-base & 19.75 & 2.61 & 12.81 \\
    PEGASUS* & 18.03 & 2.65 & 13.02 \\
    \MODEL{} & \textbf{32.97} & \textbf{11.42} & \textbf{25.41} \\
    \bottomrule
    \end{tabular}
    \caption{ROUGE score in zero-shot setting on XSum. We apply the connector to our model. \MODEL{} outperforms BART base and PEGASUS* on all metrics.}
    \label{tab:zeroshot}
\end{table}

We first need to determine the best position of mask token, as discussed in Section~\ref{sec:mask_finetuning}, where we insert the mask token before the $i$th sentence of the document, where $i=1,2,...,6$, and select the best position that achieves the highest ROUGE score on the dev collection. We report ROUGE score of all positions in \autoref{tab:mask_token_position} for the three datasets. Interestingly, we observe that the best mask token position for all datasets is before the first sentence. This agrees with the dataset generation of XSum: the summary is taken from the first sentence of the original article. For Gigaword, there is not a change after the first sentence, since the document only consists of a single sentence.

\begin{table}[t]
    \centering
    \small
    \resizebox{\columnwidth}{!}{
    \begin{tabular}{c c c | c c c  }
    \toprule
    Num Examples & RL & tok err$\downarrow$ & sent err$\downarrow$ & FactCC \\
    \midrule
    0     & 25.44 & 7.69 & 38.88 & 35.14 \\
    1     & 22.36 & 7.69 &	38.85 & 35.15 \\
    10    & 23.44 & 7.69 & 38.85 & 37.78 \\
    100   & 25.44 & \textbf{5.15} & 36.48 & \textbf{37.95} \\
    1,000 & \textbf{27.03} & 5.67 & \textbf{32.97} & 34.38 \\
    \bottomrule
    \end{tabular}
    }
    \caption{Full Result of zero-shot and few-shot experiments.}
    \label{tab:fewshot_full}
\end{table}

\section{Additional Results}

\subsection{Sentence Selection Criteria Result}
We report the full result for the sentence selection criteria in \autoref{tab:gsg_metric_full}. Surprisingly, each sentence selection criteria that uses FactCC excels in one specific factuality metric: R1+FactCC is best at FactCC, R2+FactCC is best at sentence error, and RL+FactCC is best for token error.

\subsection{Zero-shot and Few-shot}
\label{sec:fewshot_appendix}
We present additional results of the zero-shot and few-shot experiments here.

\paragraph{Zero-shot}
We first report the reference-based result of the two baseline models and \MODEL{} in \autoref{tab:zeroshot}. Due to the mismatch of pre-training and fine-tuning, we observe that both baseline models perform much worse than their result when fully trained. However, with the help of the connector, we observe 11.5 ROUGE-1 points increase for our model compared to the baseline models, and almost four times and double the score for ROUGE-2 and ROUGE-L, respectively.

\begin{table}[t]
    \centering
    \small
    \begin{tabular}{c c c c c }
    \toprule
    Model & RL & tok err$\downarrow$ & sent err$\downarrow$ & FactCC \\
    \midrule
    R1 & 29.04 & 12.31 & 60.65 & 23.93 \\
    R1+FC & 28.99 & 12.13 & 59.93 & \textbf{24.81} \\
    R2 & 29.08 & 12.12 & 59.59 &  23.67 \\
    R2+FC & 28.65 & 12.13 & \textbf{59.48} & 24.37\\
    RL & \textbf{29.23} & 12.17 & 60.08 & 23.06 \\
    RL+FC & 28.62 & \textbf{12.10} & 59.63 & 24.58 \\
    \bottomrule
    \end{tabular}
    \caption{Full result of pre-trained models with different sentence selection criteria shown in \autoref{fig:gsg_metric}. We denote the criteria with FactCC with (+FC).}
    \label{tab:gsg_metric_full}
\end{table}

\paragraph{Few-shot}
We show \MODEL{}'s full result of the few-shot experiment in \autoref{tab:fewshot_full}.

\begin{table*}[!ht]
    \centering
    \resizebox{\textwidth}{!}{
    \begin{tabular}{l c c c c  c c c c }
    \toprule
     & \multicolumn{4}{c}{WikiHow} & \multicolumn{4}{c}{Gigaword} \\
     Model & RL & tok err$\downarrow$ & sent err$\downarrow$ & FactCC & RL & tok err$\downarrow$ & sent err$\downarrow$ & FactCC \\
    \midrule
    factGSG  & \textbf{30.16} & 9.55 & 46.84 & 99.12 & 34.39 & 2.72 & 22.30 & 56.89  \\
    \midrule
    + corrector replace &  30.14 & 9.71 & 47.60 & 98.92 & 34.45 & 2.68 & 21.27 & 55.20 \\
    + corrector remove &  29.91 & 9.40 & 47.39 & 99.19 & 34.33 & 2.53 & 20.25 & 59.71 \\
    + corrector combined &  30.00 & 9.30 & 46.86 & 99.14 & 34.07 & 2.49 & 20.45 & 58.85 \\
    \midrule
    + contrastor intrinsic & 30.21 & 9.53 & 46.94 & 99.15 & \textbf{34.50} & 2.72 & 21.48 & 56.18 \\
    + contrastor extrinsic & 30.15 & 9.52 & 46.76 & 99.19 & 34.03 & 2.59 & 20.91 & 56.63 \\
    \midrule
    + contrastor + corrector & 29.91 & 8.23 & 44.59 & 99.21 & 34.38 & 2.46 & 20.04 & 58.74  \\
    \midrule
    \MODEL{} & 29.33 & \textbf{7.86}	& \textbf{42.40} & \textbf{99.41} & 34.23 & \textbf{2.30} & \textbf{19.32} & \textbf{60.02}  \\
    \bottomrule
    \end{tabular}
    }
    \vspace{-3pt}
    \caption{Fine-tuning ablation on Wikihow and Gigaword. We combine the modules by using the corrector combined and contrastor extrinsic. Results of the final model is copied from \autoref{tab:final_base}.}
    \label{tab:ablation_gw_wh}
    \vspace{-3pt}
\end{table*}
\subsection{Fine-tuning ablation on Gigaword and WikiHow}
\label{sec:ablation_gw_wh}
We report ablation of each fine-tuning components on Gigaword and Wikihow. The result can be found in \autoref{tab:ablation_gw_wh}. We observe similar trend as \autoref{tab:ablation}, where each component improves the performance. For WikiHow and Gigaword, the extrinsic method for contrastive learning perform the best. We think that this is due to the fact that the two tasks do not contain rich entities in the document, and thus require introduction of additional entities from the training corpus.

\begin{table}[!ht]
    \centering
    \resizebox{\columnwidth}{!}{
    \begin{tabular}{l c c c c  }
    \toprule
     Model & RL & tok err$\downarrow$ & sent err$\downarrow$ & FactCC \\
    \midrule
    PEGASUS* & \textbf{33.17} & 12.33 & 60.01 & 24.14 \\
    \midrule
    + corrector replace & 32.83 & 10.57 & 55.07 & 24.44 \\
    + corrector remove &  30.53 & 6.49 & 40.12 & 34.30 \\
    + corrector combined &  31.51 & 6.33 & 39.51 & 32.35 \\
    \midrule
    + contrastor intrinsic & 32.30 & 11.57 & 58.21 & 24.57 \\
    + contrastor extrinsic & 33.16 & 12.31 & 60.08 & 24.14 \\
    \midrule
    + contrastor + corrector & 31.46 & \textbf{6.22} & 39.46 & 32.39  \\
    \midrule
    PEGASUS* full &  31.49 & 6.24 & \textbf{39.37} & \textbf{32.43} \\
    \bottomrule
    \end{tabular}
    }
    \vspace{-3pt}
    \caption{Fine-tuning ablation on XSum using PEGASUS*. We combine the modules by using the corrector combined and contrastor intrinsic. We name the model with all three components as PEGASUS* full.}
    \label{tab:ablation_pegasus_star}
    \vspace{-3pt}
\end{table}
\subsection{Fine-tuning ablation using PEGASUS*}
\label{sec:ablation_pegasus_star}
We similarly perform the same ablation using the PEGASUS* model, which we present in \autoref{tab:ablation_pegasus_star}. We observe similar trend as \autoref{tab:ablation}. We note that using our pre-trained model factGSG achieves better factuality than PEGASUS* in each setting.

\section{Human Evaluation Detail}
\label{sec:human_eval_detail}
To ensure high-quality annotations, we select the workers from the United States and have more than 10,000 number of HITS approved as well as an approval rate greater than 98\%. In addition, we also create a qualification test where we rate the factuality of the selected generated summaries. Such examples include cases where some summaries hallucinate the first name of a person, which the workers should mark them as not factual. Only workers with the correct annotation can perform the actual task.

To avoid giving too much text to the workers, we select the most important sentences and replace the less relevant sentences with an ellipsis. For each of the summaries, we select the ten most relevant sentences from the document by cosine similarity of the sentence embedding using SentenceTransformer\footnote{We use the all-mpnet-base-v2  model.} \cite{reimers-gurevych-2019-sentence}. We combine and show all the selected relevant sentences from each summary. Since the summaries are similar, we see a large overlap of the relevant sentences.

We give the following prompt, which we modify from \citet{dreyer2021analyzing}:
\begin{itemize}
    \item \textit{consistency/factuality}: Please avoid using general knowledge, and only consider it in the context of the provided document. Select not consistent if facts in the summary are not supported by the document, such as cases like these:
    \begin{enumerate}
        \item The summary contradicts the information in the document. The summary might say "A fire broke out in Seattle", but a document says it broke out in Portland. Or the summary might say "the Republicans won the election", but the document indicates the Democrats won instead
        \item The summary adds (hallucinates) a fact that is not mentioned anywhere in the document. For example, the summary might say that "A fire broke out at 2 am", but the document doesn't mention the time when the fire broke out.
    \end{enumerate}
    \item \textit{Informativeness}: Please select informative if the summary expresses the main points of the document. Summary should contain relevant and important information and few unimportant details. If you select the summary to be not consistent with the document, please only consider the consistent information when evaluating this category.
\end{itemize}
                    
The order of the summary is randomly shuffled. Each task consists of three unique workers, where we take the mean as the scores for this document. The final score is the mean factuality score across all documents. The average time for each task is around 3 minutes and we pay 0.6 USD per task, hence an hourly rate of $\geq \$12$ per hour.

We use boostrap test \cite{EfroTibs93} to determine statistical significance between two summaries with $p<0.05$.

\subsection{Sample Outputs}
We include some sample outputs in \autoref{fig:sample_output} and \autoref{fig:sample_output_wh}.

\begin{figure*}[ht]
    \centering
    \resizebox{\textwidth}{!}{
    \begin{tabularx}{\linewidth}{p{\linewidth}}
    \toprule
    Example 1 (XSum) \\
    \midrule
    \textbf{Summary:} Tap water in \halent{80,000} homes in  \halent{ Lancashire} has been declared safe to drink, after the discovery of a parasite at a treatment works left residents boiling water for \halent{three weeks}. \\
    \textbf{Remove:} Tap water in homes has been declared safe to drink, after the discovery of a parasite at a treatment works left residents boiling water . \\
    \midrule
    Example 2 (XSum) \\
    \midrule
    \textbf{Summary:} \halent{Great Britain's} \halent{Becky James} won her second Olympic silver of Rio 2016 by finishing second in the women's sprint. \\
    \textbf{Remove:} Won her second Olympic silver of Rio 2016 by finishing second in the women's sprint.\\
    \midrule
    Example 3 (WikiHow) \\
    \midrule
    \textbf{Summary:} Use a "magic eraser.", Try a regular eraser. Make an alcohol solution. Apply a dry erase marker over the permanent marker. Try nail polish remover or acetone on plastic.Acetone and nail polish removers containing acetone can strip away most paints and dyes, and the dye in \halent{Sharpies} is usually weak against it. \\
    \textbf{Remove:} Use a "magic eraser.",  Try a regular eraser.  Make an alcohol solution.  Apply a dry erase marker over the permanent marker.  Try nail polish remover or acetone on plastic.Acetone and nail polish removers containing acetone can strip away most paints and dyes, and the dye is usually weak against it. \\
    \midrule
    Example 4 (WikiHow) \\
    \midrule
    \textbf{Summary:} Call \halent{911} if you think you are in danger. Cut through your denial. Do not confront your stalker. Tell other people. Keep a record of stalking activities. Get a restraining order. Be prepared to defend yourself. Change your \halent{daily} routine. Change your passwords. Use anti-virus software. Clean up your social network. Make your online materials private. Notice signs of control in your relationship. Make your break up clear. Cut off contact. Get back your keys.\\
    \textbf{Remove:} Call if you think you are in danger.  Cut through your denial.  Do not confront your stalker.  Tell other people.  Keep a record of stalking activities.  Get a restraining order.  Be prepared to defend yourself.  Change your routine.  Change your passwords.  Use anti-virus software.  Clean up your social network.  Make your online materials private.  Notice signs of control in your relationship.  Make your break up clear.  Cut off contact.  Get back your keys.\\
    \midrule
    Example 5 (Gigaword) \\
    \midrule
    \textbf{Summary:} \halent{xinhua} summary of \halent{asia-pacific} stocks news on tuesday \halent{feburary} \#\# \\
    \textbf{Remove:} summary of stocks news on tuesday \#\# \\
    \midrule
    Example 6 (Gigaword) \\
    \midrule
    \textbf{Summary:} cuba urges \halent{eu} to drop its common position \\
    \textbf{Remove:} cuba urges to drop its common position \\
    \bottomrule
    \end{tabularx}
    }
    \caption{Summaries changed using the corrector. We mark hallucinated entities in the summaries with \halent{red}.}
    \label{fig:corrector_example}
\end{figure*}

\begin{figure*}[ht]
    \centering
    \resizebox{\textwidth}{!}{
    \begin{tabularx}{\linewidth}{p{\linewidth}}
    \toprule
    XSum Example \\
    \midrule
    \textbf{Article}: The Scots started their Six Nations campaign with a thrilling first win over Ireland in four years. They beat France for the first time in 10 years last season at home, but have lost on their past nine trips to Paris. "It is a long time ago since we won there but we understand there is a reason for that," said Laidlaw. "France are a good team and it is a hard place to win. We are under no illusions this week - we need to be on top form to get something out of the game. "We'll need accuracy and we need to be physical and aggressive because the French are big men. But we need to be subtle and clever in our attack. "If we do that, and move France about and play for 80 minutes, I would back our fitness against them. But we obviously need to be in the game right to the death and not let France get any momentum." Media playback is not supported on this device France have not finished in the top half of the Six Nations since 2011, but impressed in narrow defeats by New Zealand and Australia in the autumn, and came close to a shock victory against England at Twickenham in their Six Nations opener. "They have improved massively under [coach] Guy Noves and have not been far away from any team they have played," Laidlaw noted. "They have some really dangerous strike runners throughout the team. "As much as we need to understand why we beat them last year, we need to develop that, be clever and smart and be able to score points." We need to keep doing what we are doing well, keep our skills high and if we do that again this weekend, we are going to put ourselves in a strong position. "The start of the game is massively important, whether it is our kick-off, or or us receiving. It is very important psychologically for them as it is for us. "If we can start the game playing in the right areas of the field for the first 20 minutes and not muck about too much in our own half, hopefully we will put them under pressure and upset the crowd a little bit." Media playback is not supported on this device \\
    \midrule
    \textbf{Reference}: Captain \halent{Greg} Laidlaw says Scotland must be "subtle and clever" as well as match France physically if they are to secure a first win in Paris \halent{since 1999}. \\
    \textbf{Bart-base}: Scotland head coach \halent{Greig} Laidlaw says his side are "under no illusions" when they face France at \halent{Twickenham} on \halent{Saturday}. \\
    \textbf{PEGASUS*}: Scotland coach \halent{Steve} Laidlaw has warned \halent{France} that they need to be \halent{"smart and smart"} in their attack if they are to win their \halent{first Six Nations game} of the season against the French in \halent{Paris} on \halent{Saturday}. \\
    \textbf{CLIFF}: Scotland head coach \halent{Mark} Laidlaw says his side need to be " subtle and clever" in their attack against France in \halent{Paris} on\halent{ Saturday}.\\
    \textbf{DAE}:Scotland must be " subtle and clever" in their attack when they face France at Twickenham on Saturday, says head coach \halent{Gordon} Laidlaw. \\
    \textbf{\MODEL{}}: Scotland captain Laidlaw says his side need to be "subtle and clever" in their attack if they are to beat France. \\
    \midrule
    Gigaword Example \\
    \midrule
    \textbf{Article}: the investigating magistrate and prosecutor handling the case involving the french association , zoe ' s ark , arrived tuesday in the eastern town of abeche where they met the \#\#\# children the association was attempting to fly to france , according to reports . \\
    \midrule
    \textbf{Reference}: \halent{chadian} court officials go to UNK to investigation into children 's case \\
    \textbf{Bart-base}: \halent{zoe's ark investigating magistrate} arrives in abeche\\
    \textbf{PEGASUS*}: \halent{prosecutor in zoe's ark case} arrives in abeche \\
    \textbf{CLIFF}: prosecutor meets \halent{zoe 's ark children}\\
    \textbf{DAE}: \halent{zoe's ark investigating magistrate prosecutor} arrive in abeche \\
    \textbf{\MODEL{}}: prosecutor arrives in abeche to investigate french association\\
    \bottomrule
    \end{tabularx}
    }
    \caption{Example summaries from XSum and Gigaword. Nonfactual components are marked with \halent{red}.}
    \label{fig:sample_output}
\end{figure*}

\begin{figure*}[ht]
    \centering
    \resizebox{\textwidth}{!}{
    \begin{tabularx}{\linewidth}{p{\linewidth}}
    \toprule
    WikiHow Example \\
    \midrule
    \textbf{Article}: Slide the seats all the way forward and then back to thoroughly vacuum the carpet underneath.   Start from the top and work your way down. Dust or dirt that has accumulated up top could fall down; dust or dirt that has accumulated at the bottom will rarely fall up.; , Allow it to sit for a few minutes before blotting dry with a towel. If the stain doesn't come out, repeat. After your final cleaner application, wash the area with a damp sponge and do a final blotting.   Make sure to try to get as much dampness from the fabric as possible. Any excessive dampness may promote molding and/or mildew, which does not fall in the definition of detailing a car.  Replace with a piece you've cut from a hidden place, such as underneath the seat. Use a water-resistant adhesive to keep it down...
    Warning: Always ask the owner of the car for permission before doing this step. If you want, have a sample repair that you can show the owner of the car of what the process will look like. If it's done well, this sample will be reassuring.  Apply a non-slip dressing so that the driver's feet don't slip and slide while they're trying to do important things like braking. , Use an interior dressing like Armor All to finish it off.  If you're not going to use liquids afterward, your detailing brushes should be a super-absorbant material like microfiber cloth, which picks up dust and dirt efficiently. Lightly mist some spray-on vinyl dressing onto the vent grilles to make them look brand new.  Cleaning the seats is essential for a good detail. But different seats require different methods. Note that after cleaning, you may have to vacuum out the seats or surrounding area again, as dirt will be loosened up by this process.   Cloth interiors: Interiors with nylon or other cloth can be shampooed with a wet-vac extraction machine. Cloth should be adequately dried after extraction has taken place. Leather or vinyl interiors: Interiors with leather or can be cleaned with a leather or vinyl cleaner and then hit softly with a leather brush. Cleaner can be wiped away with a microfiber cloth afterward.  If you've cleaned leather seats with product, now's the time to condition them so that the leather looks visually appealing and doesn't dry out or begin to crack.  For stronger build-up removal, use 4-ought steel wool on windows. Use plastic cleaner if the gauge cover is made of plastic...
    \\
    \midrule
    \textbf{Bart-base}: Vacuum the carpet underneath the seats. Blot the stain with a damp sponge. Replace a piece of carpet. Apply a non-slip dressing to the back of the seat. Wipe the seat down with a soft cloth. \\
    \textbf{PEGASUS*}: Vacuum the \halent{seats}. Blot the area with a damp sponge. Remove any molding and/or mildew. Apply a non-slip dressing to the area. Use a spray-on vinyl dressing to clean the vent grilles. Clean the interior of the car with a leather or vinyl cleaner. \\
    \textbf{CLIFF}: Vacuum the \halent{seats} and surrounding area. Blot the stain with a damp sponge. Remove a piece of tape from a hidden place. Cover the tape with a non-slip dressing.  , Apply an interior dressing to the back of the seat. Apply a vinyl dressing on the vent grilles. Clean the interior of the car with a microfiber cloth. Wash the seats. Condition the leather seats. Use steel wool or plastic cleaner on windows. Brush the wheel rims. Apply tire dressing. Spray degreaser on everything. Apply protectant to the tinted windows. Apply tinting to the glass. Wash and dry the car. Rinse the car thoroughly.\\
    \textbf{DAE}: Vacuum the \halent{seats} and surrounding area. Blot the area with a damp sponge. Remove a piece of wood from a hidden place. Dress the seats. Wear a non-slip dressing.  , Wipe down the vent grilles. Cover the vents.\\
    \textbf{\MODEL{}}: Vacuum the carpet under the seats. Wipe the area with a damp sponge. Apply a non-slip dressing to the seat. Wipe down the interior with a spray-on vinyl dressing. Use a leather brush to clean the interior of the car. Use steel wool on windows. Clean the interior\\
    \bottomrule
    \end{tabularx}
    }
    \caption{Example summaries from WikiHow. The article is truncated to fit the page. Nonfactual information are marked with \halent{red}.}
    \label{fig:sample_output_wh}
\end{figure*}

\end{document}